\documentclass{article}
\PassOptionsToPackage{numbers}{natbib}

\usepackage[final]{neurips_2025}

\usepackage[colorlinks=true, urlcolor=blue, citecolor=blue, linkcolor=red]{hyperref}

\usepackage[utf8]{inputenc} % allow utf-8 input
\usepackage[T1]{fontenc}    % use 8-bit T1 fonts
\usepackage{hyperref}       % hyperlinks
\usepackage{url}            % simple URL typesetting
\usepackage{booktabs}       % professional-quality tables
\usepackage{amsfonts}       % blackboard math symbols
\usepackage{nicefrac}       % compact symbols for 1/2, etc.
\usepackage{microtype}      % microtypography
\usepackage[table]{xcolor}
\usepackage{booktabs}
\usepackage{multirow}
\usepackage[pdftex]{graphicx}
\usepackage{amsmath}
\usepackage{bm}
\usepackage{tabularx}

\definecolor{aliceblue}{rgb}{0.94,0.97,1.0}
\title{\raisebox{-0.3\height}{\includegraphics[height=2.2em]{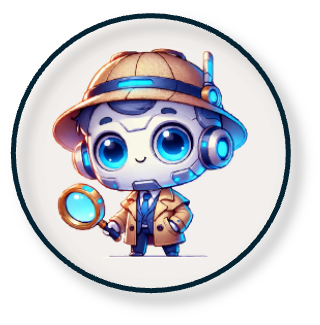}}\  Guard Me If You Know Me: \\Protecting Specific Face-Identity from Deepfakes}

\author{%
    \textbf{Kaiqing Lin}\textsuperscript{1,3*},
    \textbf{Zhiyuan Yan}\textsuperscript{2,3*}, 
    \textbf{Ke-Yue Zhang}\textsuperscript{3}, 
    \textbf{Li Hao}\textsuperscript{2}, 
    \textbf{Yue Zhou}\textsuperscript{1}, 
    \textbf{Yuzhen Lin}\textsuperscript{1}, \\
    \textbf{Weixiang Li}\textsuperscript{1}, 
    \textbf{Taiping Yao}\textsuperscript{3\dag}, 
    \textbf{Shouhong Ding}\textsuperscript{3}, 
    \textbf{Bin Li}\textsuperscript{1\dag} 
\\
\\
\textsuperscript{1}Guangdong Provincial Key Laboratory of Intelligent Information Processing,\\Shenzhen Key Laboratory of Media Security,\\and SZU-AFS Joint Innovation Center for AI Technology, Shenzhen University\\
\textsuperscript{2}School of Electronic and Computer Engineering, Peking Univerisity, \\
\textsuperscript{3}Tencent Youtu Lab \\
}

\begin{document}
\renewcommand{\thefootnote}{}  % 取消编号
\footnotetext{\textsuperscript{*} Equal contribution, \textsuperscript{\dag} Corresponding author}
\renewcommand{\thefootnote}{\arabic{footnote}}  % 恢复编号

% \renewcommand{\thefootnote}{\dag}
% \footnotetext{Corresponding authors}
% \renewcommand{\thefootnote}{\arabic{footnote}}

%Guangdong Provincial Key Laboratory of Intelligent Information Processing, \\Shenzhen Key Laboratory of Media Security, and SZU-AFS Joint Innovation Center for AI Technology,
\maketitle

\begin{abstract}
  Securing personal identity against deepfake attacks is increasingly critical in the digital age, especially for celebrities and political figures whose faces are easily accessible and frequently targeted.
  Most existing deepfake detection methods focus on general-purpose scenarios and often ignore the valuable prior knowledge of known facial identities, e.g., "VIP individuals" whose authentic facial data are already available. 
  In this paper, we propose \textbf{VIPGuard}, a unified multimodal framework designed to capture fine-grained and comprehensive facial representations of a given identity, compare them against potentially fake or similar-looking faces, and reason over these comparisons to make accurate and explainable predictions.
  Specifically, our framework consists of three main stages. First, we fine-tune a multimodal large language model (MLLM) to learn detailed and structural facial attributes. 
  Second, we perform identity-level discriminative learning to enable the model to distinguish subtle differences between highly similar faces, including real and fake variations. Finally, we introduce user-specific customization, where we model the unique characteristics of the target face identity and perform semantic reasoning via MLLM to enable personalized and explainable deepfake detection.
  Our framework shows clear advantages over previous detection works, where traditional detectors mainly rely on low-level visual cues and provide no human-understandable explanations, while other MLLM-based models often lack a detailed understanding of specific face identities.
  To facilitate the evaluation of our method, we build a comprehensive identity-aware benchmark called \textbf{VIPBench} for personalized deepfake detection, involving the latest 7 face-swapping and 7 entire face synthesis techniques for generation. 
  Extensive experiments show that our model outperforms existing methods in both detection and explanation.
  The code is available at \url{https://github.com/KQL11/VIPGuard} .
\end{abstract}

\section{Introduction}

The rapid advancement of generative AI techniques~\cite{rombach2022high,tian2024emo,xu2024vasa,yan2025gpt,roop,deepfacelab} has led to the widespread creation and dissemination of deepfake content—synthetic media where a person's identity is manipulated or replaced, often without consent. 
These techniques severely threaten personal reputation and public trust, especially for high-profile individuals such as celebrities, political figures, and public officials~\cite{bohacek2022protecting}.  
To this end, protecting one’s identity from such manipulations has become more than just a personal privacy issue—it is a pressing societal concern.

Although deepfake detection has gained attention, most existing methods are designed for general-purpose use~\cite{qian2020thinking,li2020face,shiohara2022detecting,yan2023ucf,yan2024transcending,ma2025specificity,zhou2023exposing,zhou2025breaking,yu2023learning,yu2024diffforensics,yu2025reinforced}. 
They aim to classify \textit{any} face image or video as real or fake, without considering \textit{whose} face is being targeted.
In the real world, many scenarios provide access to prior knowledge about the target identity\footnote{In this work, we refer to such individuals as "VIPs" whose identities we aim to protect.}, such as in the case of public figures or known individuals~\cite{ict_ref}. This opens a new avenue for detection: \textit{Can we leverage the known facial identity to improve both detection and explainability in personalized deepfake detection?} 
Rather than treating all faces the same, identity-aware detection methods~\cite{id1,id2,id3,ict_ref,diff_id} focus on the semantic alignment between the input image and the authentic identity. By doing so, they provide more personalized and context-aware predictions, potentially improving both detection performance and explainability.

% They often rely on low-level visual artifacts—such as blending artifacts~\cite{li2020face,shiohara2022detecting}, frequency anomalies~\cite{qian2020thinking,luo2021generalizing}, or up-sampling artifacts introduced during synthesis~\cite{tan2024rethinking,ma2025specificity}. However, such low-level signals are often subtle, forgery-specific, or easily obscured by post-processing~\cite{haliassos2022leveraging}. 
% Moreover, these detection approaches typically lack explainability. Users receive only a binary output—real or fake—without any explanation or insight into what features influenced the decision.

\begin{figure}
  \centering
  \includegraphics[width=1\linewidth]{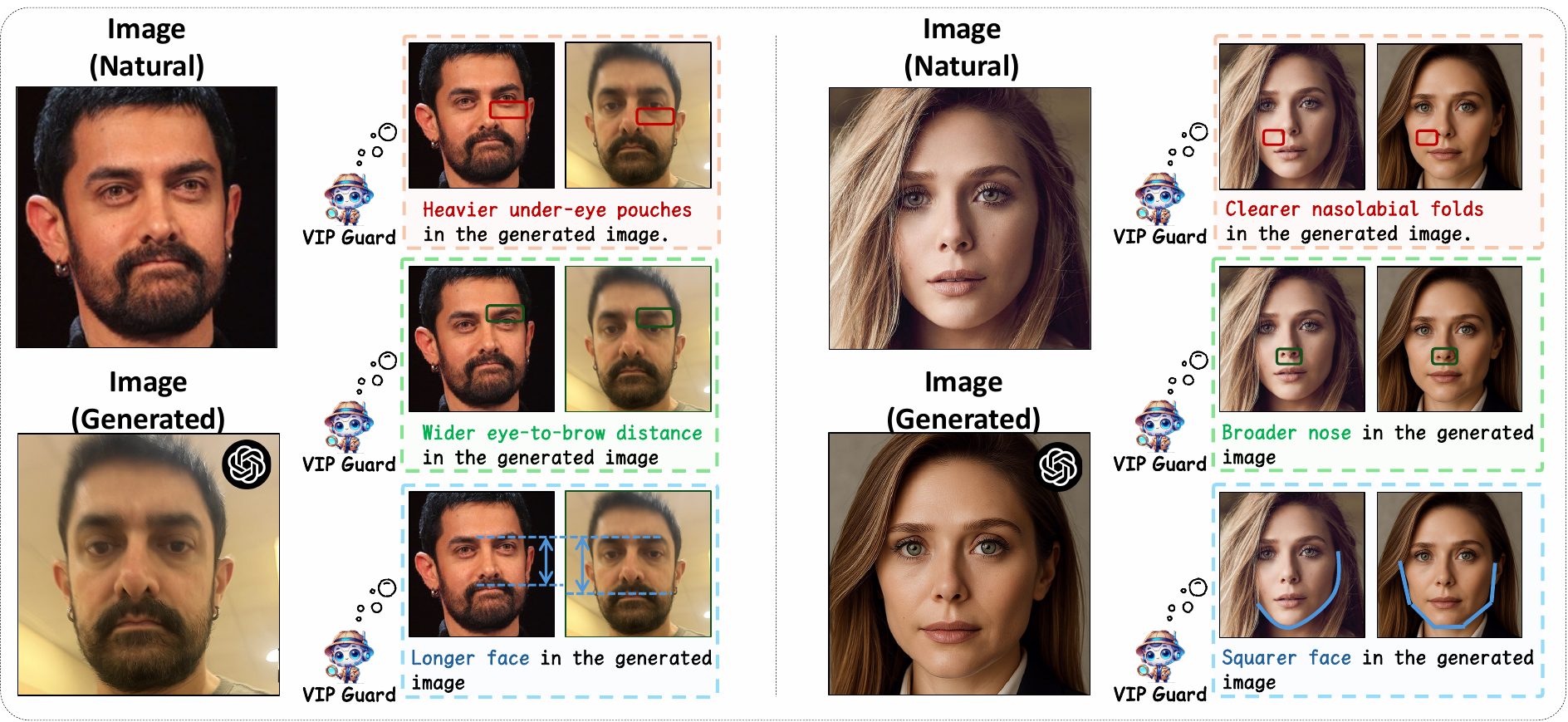}
  \label{fig:exp}
  \vspace{-4mm}
  \caption{An illustrative comparison between natural images (from LAION-Face~\cite{laion_face}) and images generated by GPT-4o~\cite{gpt4o}, showing localized inconsistencies in facial attributes, such as eye pouches and facial shapes.}
  \vspace{-1mm}
\end{figure}

However, existing identity-aware detectors~\cite{ict_ref,diff_id} fail to fully utilize the detailed identity-specific information. They primarily rely on global facial features while neglecting fine-grained semantic details—such as eye shape, facial contours, or other attributes.
% This limitation makes it difficult to catch increasingly realistic and sophisticated facial forgeries.
As shown in Figure~\ref{fig:exp}, an image generated by the latest GPT-4o~\cite{gpt4o} can appear highly realistic at first glance, but still contains subtle inconsistencies in local facial regions—for example, unusually pronounced eye bags. 
When the target identity is known and all facial details are available, leveraging these fine-grained discrepancies for detection becomes especially promising.
% Moreover, human-understandable explanations based on such subtle facial attribute differences are more intuitive and accessible for understanding.

% \begin{wrapfigure}{r}{0.4\textwidth}
  
To this end, in this paper, we propose \textbf{\textit{VIPGuard}}, a unified multimodal framework for detecting and explaining deepfakes targeting specific users. VIPGuard addresses identity-aware deepfake detection by explicitly incorporating known facial identity priors, including both global identity information and detailed structural facial attributes from the VIPs.
To achieve this, this paper, for the first time, reformulates forgery detection as a \textit{fine-grained face recognition task}, targeted at VIP identities. 
To leverage both global and local facial information, we use pre-trained face models \cite{cosface, arcface, transface} to extract global facial priors (face similarity scores) and local facial priors (facial attributes).
Leveraging these priors, we aim to enable an MLLM to perform forgery detection through semantic comparisons between suspect and authentic faces across visual and textual inputs.
% enabling reasoning beyond pixel-level artifacts.
We train VIPGuard in three stages:
(1) Fine-tuning an MLLM with a variety of facial attribute data to enhance facial understanding;
(2) Performing discriminative learning to distinguish some arbitrary identities from manipulated or similar-looking faces by reasoning;
(3) Supporting personalized detection by learning a unique and lightweight VIP token to represent each target identity for customized reasoning.

To enable robust evaluation in personalized deepfake detection, we additionally present \textbf{\textit{VIPBench}}—a \textit{comprehensive} and \textit{identity-centric} benchmark that differs fundamentally from conventional deepfake benchmarks~\cite{DeepfakeBench_YAN_NEURIPS2023,yan2024df40}. 
While existing benchmarks typically treat faces generically and overlook whose identity is being manipulated, VIPBench focuses explicitly on identity-aware scenarios, where prior knowledge of the target individual is available.
VIPBench includes 22 specific target identities and a total of 80,080 images, covering both real and forged samples. These forgeries are generated using 14 state-of-the-art methods, spanning 7 face-swapping (FS) and 7 entire-face synthesis (EFS) techniques, providing diverse manipulation types and realistic evaluation settings.
By centering evaluation around known identities and incorporating fine-grained annotations, VIPBench allows models to be assessed not only on detection accuracy but also on their ability to leverage identity-specific cues. 
% Extensive experiments demonstrate that our method, VIPGuard, achieves superior performance in both detection accuracy and explainability.

\begin{figure}
    \centering
    \label{fig:pipe}
    \includegraphics[width=1\linewidth]{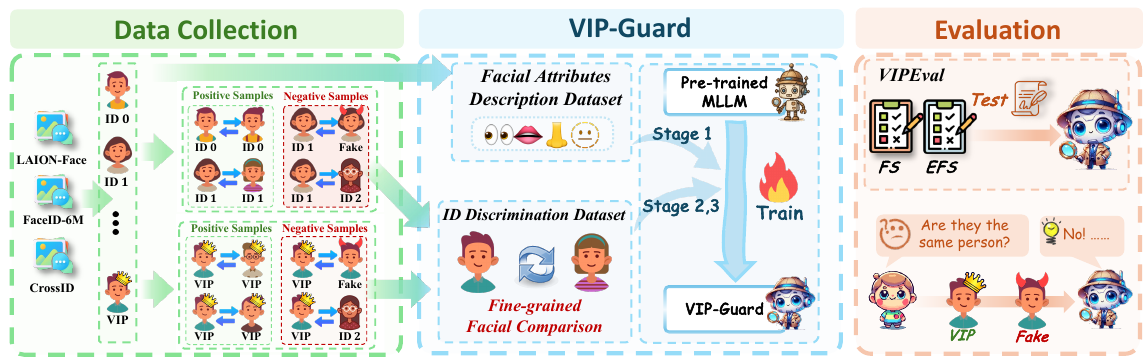}
    \caption{Overview of the data collection, VIP-Guard, and Evaluation. VIPGuard's training and inference pipeline for facial attribute understanding, identity discrimination, and VIP user customization.}
    \vspace{-3mm}
\end{figure}

Our main contributions are summarized as follows:

\begin{itemize}
    \item We introduce a new formulation for \textbf{personalized deepfake detection} that targets specific individuals, casting it as a \textit{fine-grained face recognition problem} based on both global identity features and detailed facial attributes. This formulation requires only a small number of authentic reference images per VIP user, making it practical for real-world personalized protection scenarios.

    \item We propose \textbf{VIPGuard}, a unified multimodal framework for identity-aware deepfake detection and explanation. VIPGuard incorporates pre-trained facial prior models and multimodal large language models, and follows a three-stage pipeline that extracts identity features, performs visual-text reasoning, and supports personalized detection through lightweight identity tokens.

    \item We introduce \textbf{VIPBench}, a comprehensive benchmark for evaluating identity-aware deepfake detection. It consists of \textbf{22} real-world target identities and \textbf{80,080} images generated by \textbf{14} state-of-the-art manipulation methods, enabling fine-grained and realistic assessment of personalized detection performance.
\end{itemize}

\section{Related Works}

\subsection{General Deepfake Detection}
\paragraph{Conventional Deepfake Detection}
Current deepfake detection faces significant challenges in generalization. 
To illustrate, researchers have explored a range of approaches, including data augmentation~\cite{li2018exposing,li2020face,shiohara2022detecting,chen2022self,chen2025dual}, frequency-based cues~\cite{qian2020thinking,luo2021generalizing,zhou2023exposing}, identity-aware learning~\cite{dong2023implicit,huang2023implicit}, disentanglement~\cite{yang2021learning,yan2023ucf}, reconstruction~\cite{wang2021representative,yang2025all}, and custom network designs~\cite{dang2020detection}. Data augmentation has proven especially effective for improving generalization—for example, FWA~\cite{li2018exposing} simulates warping artifacts, Face X-ray and SBI~\cite{li2020face,shiohara2022detecting} target blending boundaries, and SLADD~\cite{chen2022self} uses adversarial examples to challenge models.
Despite these advances, most traditional detectors still offer only binary outputs without human-understandable explanations, leaving users unclear about why a face is classified as fake—limiting trust and explainability.

\paragraph{Deepfake Detection via Multimodal Large Language Model}
Vision and language are two core modalities in human perception, driving growing interest in visual-language multimodal learning. 
In deepfake detection, several studies~\citep{jia2024can,shi2024shield,ye2024loki} have explored prompt-based strategies for face forgery analysis, showing that multimodal large language models (MLLMs) offer better explainability than traditional detectors. Others~\citep{foteinopoulou2025hitchhiker,xu2025fakeshield,xu2025avatarshield,li2024forgerygpt,wen2025spot,kang2025legion,zhang2024common,he2025vlforgery} have investigated different MLLMs for explainable detection, while \cite{li2024fakebench} introduced a labeled multimodal dataset to support fine-tuning. 
Moreover, $\mathcal{X}^2$-DFD~\citep{chen2024x2dfd} further investigates hybrid frameworks that integrate conventional visual models with MLLMs.
However, most of these methods are designed for general-purpose detection and overlook the valuable identity-specific information available in many real-world scenarios.

\subsection{Personalized Deepfake Detection for Specific Identity Protection}
To detect identity inconsistencies in forged faces, prior works~\cite{id1,id2,id3,ict_ref,cozzolino2021id,ming2025critical} use reference images for personalized deepfake detection. 
For example, ICT-Ref~\cite{ict_ref} employs a transformer to detect mismatches between inner- and outer-face regions, while DiffID~\cite{diff_id} uses reconstruction-based identity distances to identify fakes.
However, these methods mainly rely on global identity features and overlook deeper, user-specific information. This limits their robustness to distribution shifts and prevents them from offering human-understandable explanations.

% To address these gaps, we propose a user-specific deepfake detection framework called VIPGuard that leverages pre-trained MLLMs to reason about forgeries using both global identity priors and VIP-specific facial attributes. Our approach also generates human-understandable explanations to enhance transparency. 还是别加这句，空间不够（zhiyuan）
% Notably, a concurrent independent work~\cite{xu2025identity} also explores user-specific detection but focuses on general semantic anomalies (e.g., clothing, eyewear) rather than detailed facial attributes. In contrast, our approach provides a finer-grained analysis by explicitly modeling facial attribute inconsistencies, making it more precise and effective for identity-specific deepfake detection.
% As its focus differs from ours, we do not include it in our comparisons.

\section{\textit{VIPBench}: A New Benchmark for Personalized Deepfake Detection}
\vspace{-2mm}
To promote the training and evaluation of personalized deepfake detection, we build a comprehensive identity-aware benchmark called VIPBench for personalized deepfake detection.
% This dataset can
% We define facial forgery detection for specific users as a fine-grained form of facial recognition.
% To equip MLLMs with user-specific facial recognition capabilities for forgery detection and interpretable explanation generation, we developed the Facial Attributes Description Dataset and Identity Discrimination Dataset, shown in Figure \ref{fig:data_pipe}. 
The training set of VIPBench progressively fine-tunes MLLMs, advancing from basic facial attribute recognition to fine-grained identity inconsistency detection.
Moreover, we introduce a new evaluation dataset for identity-aware deepfake detection, a setting that currently lacks sufficient evaluation resources, to assess the effectiveness of different methods.
We obtain all facial images from open-sourced datasets, including LAION-Face~\cite{laion_face}, CrossFaceID~\cite{crossfaceid}, and FaceID-6M~\cite{faceid6m}, subjected to some preprocessing (details in supplementary materials).
The main idea of the dataset construction pipeline is described below, while complete details are provided in the supplementary materials.

\begin{figure}
  \centering
  \includegraphics[width=1\linewidth]{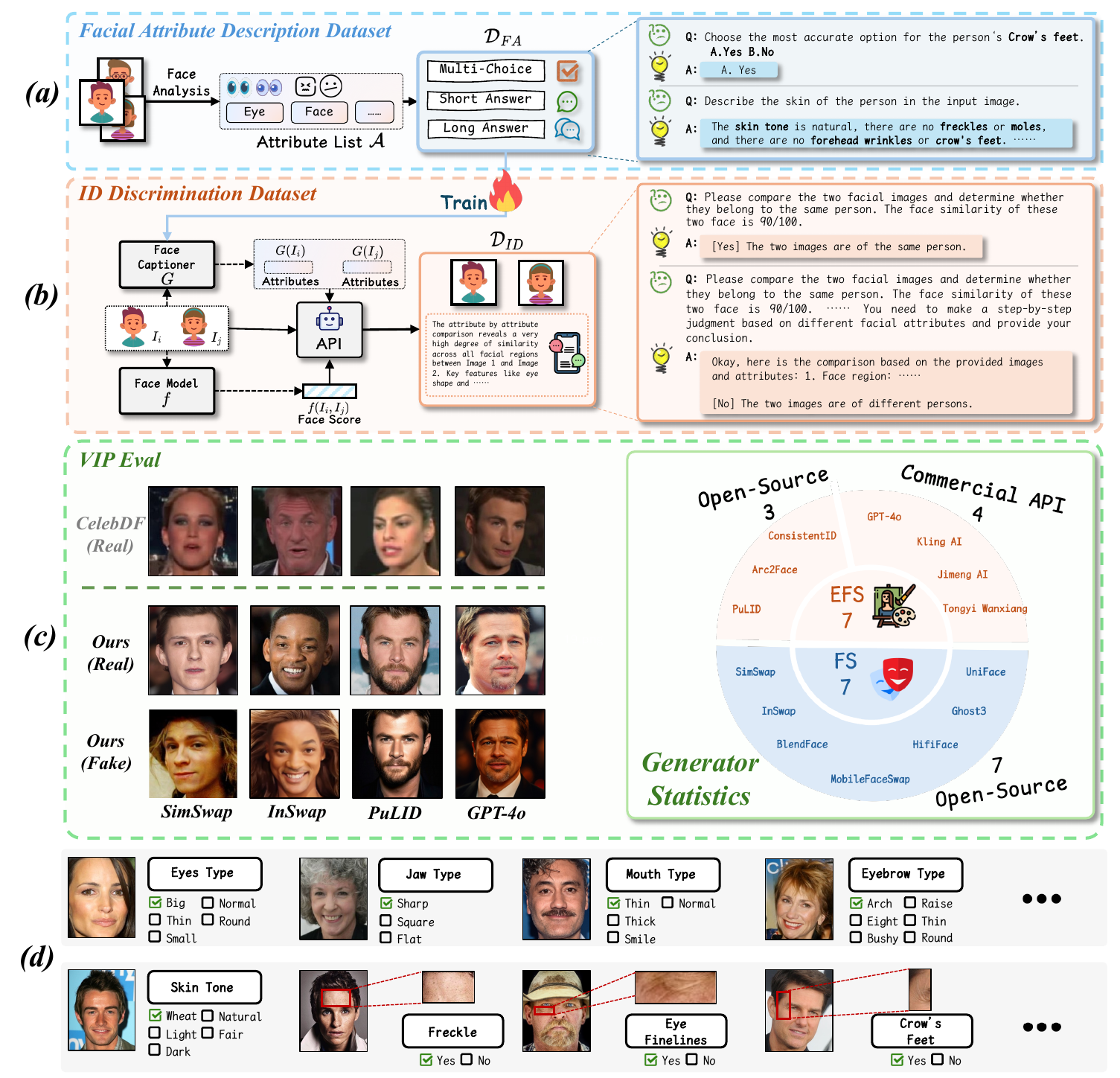}
\caption{Illustration of the proposed VIPBench, which includes three personalized datasets, (a) Facial Attributes Description Dataset $\mathcal{D}_{FA}$, (b) Identity Discrimination Dataset $\mathcal{D}_{ID}$, and (c) VIPEval $\mathcal{D}_{Eval}$. (d) Some examples of the facial attributes used in the $\mathcal{D}_{FA}$ are also illustrated here, while the full set is available in the supplementary material. The real images shown in (c) are from CelebDF~\cite{laion_face} and VIPBench, while the fake ones are generated using multiple models. All images in (d) are sourced from LAION-Face~\citep{laion_face}.}
  \label{fig:data_pipe}
  \vspace{-5mm}
\end{figure}

\paragraph{Facial Attributes Description Dataset}
\label{subsec:fa_data}

We present the Facial Attribute Description Dataset—a multimodal dataset composed of high-resolution facial images paired with rich facial attribute descriptions. The dataset ($\mathcal{D}_{FA}$) is collected about 30,000 high-resolution (more than $1024\times 1024$) images from LAION-Face, to facilitate foundational facial understanding in MLLMs.
In addition, an MLLM specialized in facial analysis can act as a captioner to generate descriptive facial attribute information.
As illustrated in Figure \ref{fig:data_pipe}, we obtained detailed attributes (e.g., face shape, skin condition) via MegVii's official API\footnote{https://www.faceplusplus.com.cn/} and subsequently refined them by human experts.
Figure \ref{fig:data_pipe} (d) provides some examples of these facial attributes, which were then transformed into diverse VQA formats (multiple-choice, short/long answer).
The process can be formulated as Eq. \ref{eq:dfa}
\begin{equation}
\begin{aligned}
\mathcal{D}_{FA} = \left\{ \left(I_i, \text{VQA}(a_{i1}, a_{i2}, \ldots, a_{ik})\right) \;\middle|\; (a_{i1}, \ldots, a_{ik}) \in \mathcal{A}^k,\; i = 1, \ldots, N \right\},
\end{aligned}
\label{eq:dfa}
\end{equation}
where $\mathcal{A} = F(I)$ denotes the set of extracted facial attributes, $F$ is the API, and $I$ is the facial image. The dataset $\mathcal{D}_{FA}$ comprises VQA instances generated from all $k$-tuples of attributes drawn from $\mathcal{A}$.
This dataset provides rich supervision for training an MLLM to understand fundamental facial characteristics.
The other details about $\mathcal{D}_{FA}$ are shown in the supplementary material.
% To establish foundational facial understanding in VLMs, we constructed . 
% We collected about 30,000 high-resolution (more than $1024\times 1024$) images from LAION-Face, extracting facial regions. Detailed attributes (e.g., face shape, skin condition) were obtained via MegVii's commercial API and subsequently refined by human experts. These attributes were transformed into diverse VQA formats (multiple-choice, short/long answer), with long answers synthesized using GPT-4o prompts. 
% This dataset provides robust supervision for teaching VLMs fundamental facial characteristics.

\paragraph{Identity Discrimination Dataset}
\label{subsec:id_data}
By reformulating personalized deepfake detection as a target-face-centric and fine-grained face recognition problem, we constructed the Identity Discrimination Dataset $\mathcal{D}_{ID}$.
This dataset, comprising $\mathcal{D}_{ID}^{general}$ and $\mathcal{D}_{ID}^{vip}$, includes facial images and corresponding annotations intended for reasoning about fine-grained identity discrimination.
% We created an image-name pool $\mathcal{J}$ by extracting character names from image captions via the DeepSeek API~\cite{deepseek}, facilitating the retrieval of image pairs representing both positive pairs and negative pairs. 
Specifically, $\mathcal{D}_{ID}$ comprises facial image pairs: positive (same-identity, real-real) and negative (all others). 
As shown in the `Data Collection' part of Figure ~\ref{fig:pipe}, we built facial pairs centered on VIP users in $D_{ID}^{vip}$, while facial pairs in $\mathcal{D}_{ID}^{{general}}$ were built using arbitrary identities.
Given the high resemblance of forged faces to genuine ones, which is challenge for conventional face recognition, we augmented negative samples using SimSwap~\cite{simswap} (face swapping) and Arc2Face~\cite{arc2face} (entire face synthesis). These negative pairs include different-identity real-real pairs and real-versus-forgery pairs, with these two categories balanced roughly 1:1. 
% We then generated identity-comparative VQA reasoning annotations for these pairs using Gemini 2.5 Pro~\cite{Gemini}, based on facial priors (prompts in supplementary).
As demonstrated in Figure ~\ref{fig:data_pipe} (b), for each image pair, we first generated a facial attribute list using our fine-tuned Qwen2.5VL-7B model~\cite{bai2025qwen2} $G$, trained on $\mathcal{D}_{FA}$. 
Then, we acquired the facial similarity scores by a pre-trained facial recognition model $f$.
Finally, we utilized a commercial model (Gemini 2.5 Pro\footnote{Gemini API version in use: 2.5-pro-exp-03-25.}~\cite{Gemini} was selected in this paper), denoted as API in Figure ~\ref{fig:data_pipe} (b), to form the training data.
Specifically, \textit{the API was then strictly required to reason identity discrepancies between image pairs based solely on the provided facial attributes $G(I_i)$ and $G(I_j)$ and similarity scores $f(I_i, I_j)$, ensuring that the analysis was grounded in real features and not influenced by hallucinated content.}
% Besides, we also provided the prior information to Gemini for the negative pairs by explicit prompts (e.g., `\texttt{Although the faces may appear similar, they are not the same person.}'). 
The above process can be formulated as 
\begin{equation}
\begin{split}
\text{Prompt} &= \text{API}\left(Q, f(I_i, I_j), I_i, I_j, G(I_i), G(I_j)\right) \\
\mathcal{D}_{ID} &= \left\{I_i, I_j, \text{VQA}\left(I_i, I_j, \text{Prompt}\right) \;\middle|\; I_i, I_j \in \mathcal{J} \right\},
\end{split}
\label{eq:did}
\end{equation}
where $I_i$ and $I_j$ are facial images, $Q$ is the prompt inputted into Gemini, and $\mathcal{J}$ denotes a pre-constructed pool of image-name pairs (see supplemental material for details).
% We also constructed a training set $\mathcal{D}^{vip}_{ID}$ for VIP users in the same manner, while ensuring non-overlapping identities to prevent identity leakage.

\paragraph{VIPEval (User-Specific Evaluation Dataset):}
\label{subsec:vip_data}
We introduce a user-specific evaluation dataset called \textbf{VIPEval} for assessing user-specific forgery detection performance.
Conventional datasets~\cite{li2019celeb,rossler2019faceforensics++} typically lack a sufficient number of high-resolution real images per individual, along with corresponding variations across different face forgery methods, which makes it challenging to evaluate personalized forgery detection approaches.
The dataset includes images from rich sources and various resolutions with diverse generation methods, making it more representative of real-world conditions.
To address this limitation, we carefully selected 22 unique identities from the previously described image-name pool $\mathcal{J}$, ensuring no overlap with the identities in the $\mathcal{D}_{ID}^{general}$. 
For each identity, 40-60 real images were collected, 20 reserved for testing in this benchmark $\mathcal{D}_{Eval}$, and the remainder used to construct $\mathcal{D}_{ID}^{vip}$.
% Notably, to prevent identity leakage in face swapping detection, we deliberately avoid using the 22 benchmark identities when constructing negative training pairs involving each other.
The benchmark includes reserved real images and a comprehensive set of forged counterparts. For each of the 22 test identities, 420 images are generated per method using seven distinct face-swapping (FS) techniques~\cite{simswap,blendface,inswapper,uniface, xu2022mobilefaceswap,wang2021hififace,ghost3}. Additionally, three open-source entire-face synthesis (EFS) methods~\cite{arc2face,huang2024consistentid,guo2024pulid} generate 10 images per real test image by varying random seeds or prompts, resulting in 200 images per identity. Furthermore, four commercial API-based EFS methods~\cite{gpt4o, jimengai, TongYiWanXiang, kling} produce 20 images per identity.
In total, the dataset comprises 80,080 images.

\section{\textit{VIPGuard}: A Multimodal Framework for Personalized Deepfake Detection}
\vspace{-2mm}
\subsection{Problem Formulation and Comparison with Prior Works}
\vspace{-2mm}
We consider a personalized deepfake detection scenario in which the detector has access to several authentic images of the target user (e.g., a celebrity, hereafter referred to as a VIP), along with a collection of real facial images from other unrelated individuals. 
These authentic images serve as prior knowledge for the detector to better recognize and protect the target individual. 
The objective is to model user-specific facial characteristics to identify suspicious images and protect the VIP’s identity from forgery attacks.
For forgeries, such as face-swapped images, we impose a realistic constraint that the source identities used for manipulation (denoted as $ID^{source}$) are unseen by the detector during training. This assumption is closer to the real-world scenario, where attackers can use arbitrary faces that are not available to the defender.
This setting contrasts with identity-aware detection methods~\cite{ict_ref, id2, ming2025critical}, which assume that both the target identity ($ID^{vip}$) and source identities ($ID^{source}$) are known during training—i.e., all relevant faces are included in the reference set $\{ID^{vip}, ID^{source}, ID^{others}\}$. Such a closed-world assumption simplifies the problem but is rarely realistic in practical applications.
In our formulation, the detector only has access to $\{ID^{vip}, ID^{others}\}$ during training, while $ID^{source}$ remains unknown. 
This difference introduces a more challenging yet practical problem setting, emphasizing the need for generalization to unseen source identities and better alignment with real-world deployment scenarios.
\vspace{-3mm}

\subsection{Method Overview}
% We address the challenge of protecting specific users (e.g., celebrities) from forgery attacks by leveraging a limited set of authentic facial images.
\vspace{-2mm}
To address the challenge, we propose a framework, VIPGuard, which develops an MLLM capable of identifying suspicious images of a specific user, while providing human-understandable explanations based on the user's unique facial attributes.
We reformulate personalized deepfake detection as a fine-grained face recognition problem centered on the protected target, where forgeries are detected through the MLLM’s reasoning over global identity features and detailed facial attributes.
To equip the model with this capability, we start from a pre-trained MLLM (Qwen-2.5-VL-7B~\cite{bai2025qwen2}) and progressively fine-tune it through a three-stage process: Face Attribute Learning, Identity Discrimination, and User-Specific Customization.
We describe each stage in detail below.
% In Stage 1, we adapt the model to the facial domain using $\mathbf{D}_{FA}$ (see Section~\ref{subsec:fa_data}), which contains fine-grained and detailed facial attribute annotations.
% In Stage 2, we enhance the model’s capability for fine-grained identity discrimination across arbitrary facial images by training it on $\mathbf{D}_{ID}^{general}$ (see Section~\ref{subsec:id_data}).
% In Stage 3, we further train identity-specific tokens on $\mathbf{D}_{ID}^{vip}$ (see Section~\ref{subsec:id_data}) to enable personalized forgery detection for a given VIP user.
\vspace{-2mm}

\subsection{Three-Stages Training of VIPGuard}
\vspace{-2mm}
\paragraph{Stage 1: Face Attributes Learning}
It is crucial first to enhance the model’s capability to recognize and utilize fine-grained facial attributes, as naive MLLMs inherently lack a sufficient understanding of human facial features for effective VIP identity protection. 
To this end, we fine-tune the pre-trained MLLM on $\mathcal{D}_{FA}$, which contains a large number of samples for facial attribute recognition (see Section~\ref{subsec:fa_data}).
We integrate LoRA~\cite{lora} modules into the pre-trained MLLM and perform supervised fine-tuning using an autoregressive loss, as defined in Eq.~\ref{eq:stage1}:
\begin{equation}
\centering
\label{eq:stage1}
L(\boldsymbol{\theta}) = - \sum_{i=1}^N \log \left[p_{\boldsymbol{\theta}}(x_i \mid x_{<i}, E_V(I))\right],
\end{equation}
where $\boldsymbol{\theta}$ denotes the parameters of both the inserted LoRA modules and the original MLLM, $N$ is the length of the output prompt, $x_i$ is the $i$-th token to be predicted while $x_{<i}$ are the previous tokens, $E_V$ represents the visual encoder of the MLLM, and $I$ is the input image.
After training, the MLLM can understand and then recognize facial attributes.

\begin{figure}
  \centering
  \includegraphics[width=1.0\linewidth]{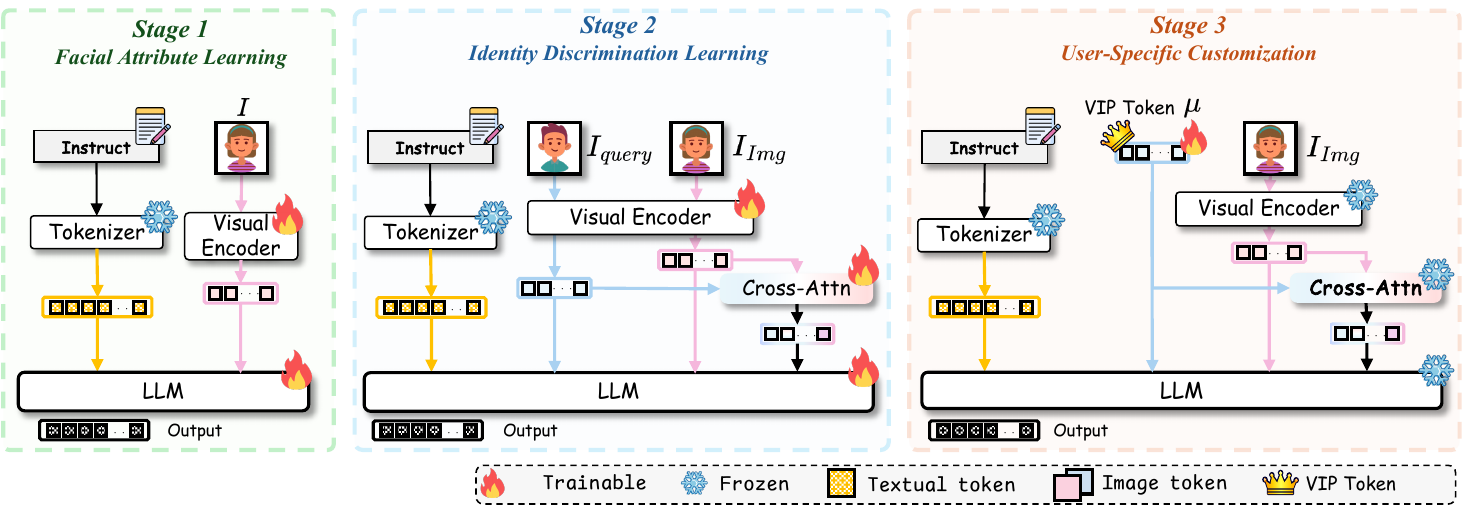}
  \caption{Illustration of the three stages of training the proposed VIPGuard framework.}
  \label{fig:method}
\end{figure}

\vspace{-2mm}
\paragraph{Stage 2: Identity Discrimination Learning}
As illustrated in Figure~\ref{fig:exp}, although the fake face appears realistic at first glance, subtle discrepancies remain in its detailed facial attributes.
Therefore, we reformulate personalized deepfake detection as a fine-grained face recognition problem centered on the protected target, where forgeries are detected through the MLLM’s reasoning over global identity features and detailed facial attributes.
In this stage, we further fine-tune the MLLM on $\mathcal{D}_{ID}^{{general}}$ for fine-grained facial recognition across face pairs—\textit{each consisting of an arbitrary query face} $I_{{query}}$ \textit{and an arbitrary input face} $I_{{Img}}$—thereby equipping it with a foundational ability to identity reason.
$\mathcal{D}_{ID}^{\text{general}}$ contains a large number of positive (same-identity) and negative (including both different-identity and forgery) face pairs, each annotated with VQA-style reasoning questions and answers. 
These annotations support fine-grained identity discrimination by reasoning over both global and local facial prior.
Here, we employ face recognition models~\cite{cosface, arcface, transface} to provide face similarity scores, serving as a global facial prior, while the local facial prior—i.e., knowledge of facial attributes—has already been incorporated into the fine-tuned MLLM during Stage 1.
Specifically, as shown in Figure ~\ref{fig:method}, the query $I_{query}$ and the input image $I_{Img}$ are first inputted into the visual encoder $E_V$ of our MLLM to obtain the visual token sequences $f_{query}$ and $f_{Img}$.
Pre-trained MLLMs typically lack task-specific optimization for facial recognition, rendering the vanilla visual features $f_{query}$ and $f_{Img}$ suboptimal for facial identity discrimination. 
Hence, we introduce a Cross-Attention (Cross-Attn) module~\cite{crossattn} to capture fine-grained differences in visual tokens between distinct facial images, which is formulated as follows
\begin{equation}
    \centering
    g = \mathrm{softmax}\!\left(\frac{QK^{\top}}{\sqrt{d_K}}\right)V, 
    \quad \text{where } Q = f_{\text{query}},\; K = V = f_{\text{Img}}.
\end{equation}
% As illustrated in Figure ~\ref{fig:method}, a cross-attention layer followed by a feed-forward network (FFN) is incorporated to explicitly model the interactions between $f_{ref}$ and $f_{query}$, thereby enhancing the representation of identity-related features. 
% The extraction of interaction features $g$ is defined in Eq.~\ref{eq:ca}.
% \begin{equation}
%     \centering
%     \label{eq:ca}
%     g = FFN(softmax(\frac{Q(f_{ref}) \cdot K(f_{query})^T}{\sqrt{d}}) \cdot V(f_{query})),
% \end{equation}
% where $Q$, $K$, and $V$ are fully connected layers (with dimension $d\times d$ in cross-attention mechanism, corresponding to the query, key, and value, respectively. 
% This module can capture the discrepancy between two input face representations, enhancing the model's ability to discriminate facial identities. 
We optimize the MLLM by an autoregressive loss shown as Eq. ~\ref{eq:stage2} below.
\begin{equation}
  \centering
  \label{eq:stage2}
  L(\boldsymbol{{\theta}}) = - \sum_{i=1}^N \log \left[p_{\boldsymbol{{\theta}}}(x_i \mid x_{<i}, f_{query}, f_{Img}, g)\right].
\end{equation}
After training, the model can perform fine-grained facial recognition between any two faces, while also providing detailed explanations of their differences.
% We aim to customize an MLLM for a specific VIP user to not only detect forged facial images but also generate personalized explanations based on that user’s facial characteristics. Providing user-centric, identity-grounded textual explanations enhances the interpretability of the detection process.
% However, most existing open-source pre-trained MLLMs typically lack strong facial identity discrimination capabilities. This challenge is further compounded by FS and EFS techniques, which produce synthetic faces that closely resemble real ones—differing only in subtle, fine-grained semantic details.
% To tackle this challenge, we further fine-tune the MLLM on $\mathcal{D}_{ID}$ for the general identity discrimination task, equipping it with the capability to recognize identities based on detailed facial attribute representations.

\vspace{-2mm}
\paragraph{Stage 3: User-Specific Customization}
Upon completing Stage 2, the MLLM acquires a basic capability for fine-grained facial identity comparison with reasoning.
In Stage 3, we further introduce a user-specific customization to enable personalized forgery detection for a given VIP user.
To this end, we construct a dataset $\mathcal{D}_{ID}^{{vip}}$ with the same structure as $\mathcal{D}_{ID}^{{general}}$, centered on the VIP’s identity, allowing the model to identify suspicious images by reasoning based on the VIP user's facial prior.
Motivated by the Yo'LLaVA~\cite{nguyen2024yo}, to facilitate lightweight deployment, we incorporate a learnable VIP token trained on $\mathcal{D}_{ID}^{vip}$, denoted as $\boldsymbol{\mu}$, which encodes identity-specific features of the VIP user.
During this stage, the parameters of the MLLM from Stage 2 are frozen, and only the VIP token is trained on $\mathcal{D}_{ID}^{{vip}}$, thereby refining the model’s ability to perceive and distinguish the target VIP user.
As illustrated in Figure \ref{fig:method}, the learnable VIP token $\boldsymbol{\mu}$, which is a vector of size $32 \times d$, is employed to substitute the visual feature representation $f_{query}$ of the query image $I_{query}$.
Formally, the prediction procedure can be described as follows
\begin{equation}
    \centering
    g = \mathrm{softmax}\!\left(\frac{QK^{\top}}{\sqrt{d_K}}\right)V, 
    \quad \text{where } Q = \boldsymbol{\mu},\; K = V = f_{\text{Img}}.
\end{equation}
\vspace{-2mm}
\begin{equation}
  \centering
  \label{eq:stage3}
  L(\boldsymbol{\mu}) = - \sum^N_{i=1} log [p_\theta(x_i|x_{<i}, \boldsymbol{\mu}, f_{query}, g)].
\end{equation}
Notably, the query images in $\mathcal{D}_{ID}^{{vip}}$ are not used; only the input image $I_{Img}$ and the corresponding reasoning annotation are fed into the MLLM.
After training, the MLLM can perform personalized forgery detection for the VIP user by determining whether the identity of the input image $I_{{Img}}$ belongs to the target individual and providing a user-centric explanation.
Furthermore, during inference, we can substitute the corresponding VIP token $\boldsymbol{\mu}$ for each user, enabling efficient and accurate detection of malicious facial forgeries without requiring any model retraining. This design supports lightweight and scalable deployment.

\vspace{-2mm}
\section{Experiments}
\vspace{-2mm}
In this section, we present comprehensive experiments to evaluate the effectiveness of our method. For general deepfake detectors, we used models trained on standard deepfake datasets. For ID-aware detectors, we assumed access to 20-30 real images of each VIP user. Detailed experimental settings are provided in the supplementary material.
% \subsection{Experimental Settings}
% \paragraph{Dataset}
% \paragraph{Implementation details}
% We adopt the pre-trained Qwen-2.5-VL model as the backbone.
% The model was optimized using the Adam optimizer with a cosine learning rate decay schedule, starting from an initial learning rate of 3e-5.
% To accommodate GPU memory limitations, the effective batch size was maintained at 72 for both Stage 1 and Stage 2 by applying gradient accumulation.
% In Stage 3, the effective batch size was reduced to 8 and the initial learning rate was set to 1.
% All training was performed using mixed-precision computation within the open-source Swift framework.
% The model was trained for 2 epochs in Stage 1, and for 1 epoch each in Stage 2 and Stage 3.
\vspace{-2mm}
\paragraph{Evaluation Metrics}
For evaluation, we report three standard metrics: Area Under the Curve (AUC), Equal Error Rate (EER), and Accuracy (ACC). AUC measures the model’s ability to distinguish between positive and negative classes across all thresholds. EER represents the point where the false acceptance rate is equal to the false rejection rate. ACC denotes the proportion of correct predictions. We use AUC and EER to compare our method following prior work~\cite{yan2024df40,DeepfakeBench_YAN_NEURIPS2023}. 
Moreover, ACC is used to compare with API-based commercial generators (e.g., GPT-4o\footnote{GPT4o API version in use: GPT-4o-2024-08-06}~\cite{gpt4o}).

\vspace{-2mm}
% \subsection{Evaluations on the proposed VIPEval and Other Existing Benchmarks}
% \vspace{-2mm}
\paragraph{Comparison with Deepfake Detection Methods on VIPEval}
We evaluated general deepfake detectors, ID-aware detectors, and our proposed method on VIPEval, reporting results for face swapping (Table ~\ref{tab:faceswap}) detection and entire faces synthesis detection (Table ~\ref{tab:efs}). 
Our method significantly improves the detection of all facial forgery types. 
General deepfake detectors (e.g., Effort~\cite{yan2024effort}) excel at face swapping detection but fail to detect more realistic forgeries from commercial APIs due to unseen artifacts. 
ID-aware detectors, unlike general detectors reliant on low-level cues, leverage identity-related semantic consistency for robust detection across diverse generation techniques and data. 
However, existing ID-aware methods (e.g., DiffID, ICT-Ref) struggle with fully synthesized faces due to their reliance on global facial features, a limitation amplified by advancing forgery techniques.
Conversely, VIP-Guard captures identity-specific cues by jointly leveraging global facial representations and local attribute analysis, enabling robust protection of VIP users with few real photos across diverse forgery methods.

\begin{table}[h]
  \centering
  \tiny
  \caption{Evaluation of generalization performance (AUC (\%) / EER (\%)) for face swapping detection on VIPEval.}
  \label{tab:faceswap}
  \begin{tabular}{l|ccccccc}
    \toprule
    \textbf{Methods} &{BlendFace~\cite{blendface}} &Ghost3~\cite{ghost3} &HifiFace~\cite{wang2021hififace} &InSwap~\cite{inswapper} &MobileSwap~\cite{xu2022mobilefaceswap} &SimSwap~\cite{simswap} &UniFace~\cite{uniface}  \\
    \midrule
    Xception~\cite{chollet2017xception} & 53.89 / 46.59 & 61.08 / 42.27 & 71.70 / 34.09 & 64.79 / 38.64 & 98.12 / 7.50 & 64.91 / 38.41 & 59.91 / 42.50  \\ 
    EfficientNet~\cite{tan2019efficientnet} & 33.94 / 60.91 & 43.52 / 54.32 & 61.22 / 42.95 &37.34 / 58.64 & 89.67 / 64.45 & 56.55 / 45.00 & 46.53 / 50.23 \\
    UCF~\cite{yan2023ucf} & 64.31 / 38.18 & 65.92 / 35.68 & 65.62 / 36.82 & 66.02 / 34.32 & 87.33 / 20.00 & 63.11 / 36.82 & 67.25 / 34.32 \\ 
    ProDet~\cite{cheng2024can} & 59.84 / 42.95 & 53.18 / 47.73 & 56.48 / 46.36 & 35.91 / 59.32 & 73.01 / 34.32 & 56.27 / 45.68 & 49.34 / 50.00 \\
    RECCE~\cite{cao2022end} & 56.11 / 45.00 & 61.53 / 39.54 & 56.93 / 45.91 & 58.85 / 43.41 & 88.40 / 19.32 & 60.05 / 42.95 & 63.01 / 39.10 \\
    CDFA~\cite{lin2024fake} &59.86 / 45.23 &70.50 / 36.82 &85.43 / 24.09 & 70.92 / 36.59 & 98.75 / 5.00 & 71.22 / 35.91 & 73.30 / 34.77 \\
    RepDFD~\cite{lin2025standing} &70.34 / 35.00 & 78.76 / 28.18 & 80.09 / 27.50 & 71.66 / 34.32 & 95.76 / 10.45 & 79.40 / 27.95 & 82.36 / 25.45 \\ 
    Effort~\cite{yan2024effort} & 91.87 / 17.61 & 95.28 / 13.07 & 96.60 / 10.23 & 85.53 / 23.86 & 96.23 / 13.07 & 97.16 / 8.52 & 92.48 / 13.64  \\
    \midrule 
    DiffID~\cite{diff_id} & 83.33 / 24.47 & 66.02 / 38.38 & 87.23 / 20.33 & 66.22 / 38.06 & 75.71 / 30.89 & 72.03 / 33.91 & 83.30 / 24.40 \\
    ICT-Ref~\cite{ict_ref} & 88.67 / 13.37 & 86.52 / 15.51 & 85.90 / 14.05 & 84.34 / 17.87 &87.45 / 13.52 &86.57 / 14.65 &82.73 / 19.04  \\
    \midrule
    \rowcolor{aliceblue}
    VIPGuard &\textbf{99.48} / \textbf{0.81} & \textbf{97.97} / \textbf{4.39} & \textbf{99.63} / \textbf{0.61} & \textbf{96.40} / \textbf{8.24} & \textbf{99.55} / \textbf{1.01} & \textbf{99.43} / \textbf{1.02} & \textbf{99.69} / \textbf{0.26}  \\
    \bottomrule
  \end{tabular}
\end{table}

\begin{table}
  \centering
  \tiny
  \caption{Evaluation of generalization performance (AUC (\%) / EER (\%)) for entire face synthesis detection on VIPEval.}
  \label{tab:efs}
  \begin{tabular}{l|ccccccc}
    \toprule
    \multirow{2}{*}{\textbf{Method}} &\multicolumn{3}{c}{\textbf{Open-Source}} &\multicolumn{4}{c}{\textbf{Commercial-API}} \\
     &ConsistentID~\cite{huang2024consistentid} &Arc2Face~\cite{arc2face} &PuLID~\cite{guo2024pulid} &GPT-4o~\cite{gpt4o} &Jimeng AI~\cite{jimengai} &TongYi~\cite{TongYiWanXiang} &Kling AI~\cite{kling} \\
    \midrule
    Xception & 42.02 / 54.77 & 51.87 / 48.86 & 59.23 / 44.09 & 58.13 / 46.36 & 57.77 / 44.55 &34.36 / 62.05 &44.34 / 54.32 \\ 
    EfficientNet &33.81 / 61.14 &44.96 / 50.91 &47.19 / 50.23 &75.35 / 28.86 &55.04 / 45.00 &41.15 / 55.68 &42.40 / 54.32 \\
    UCF~\cite{yan2023ucf} & 62.16 / 40.91 & 56.62 / 46.36 & 54.16 / 46.36 & 59.06 / 42.73 & 71.08 / 32.78 & 82.38 / 22.73 & 63.31 / 40.23  \\
    ProDet~\cite{cheng2024can} & 63.68 / 37.95 & 59.62 / 40.91 & 67.02 / 36.82 & 59.23 / 40.91 & 59.71 / 42.73 & 89.80 / 18.41 & 72.53 / 32.73 \\
    RECCE~\cite{cao2022end} & 68.56 / 36.82 & 57.79 / 47.72 & 63.85 / 40.00 & 83.63 / 22.73 & 69.00 / 35.23 & 97.00 / 7.50 & 70.94 / 37.72 \\
    CDFA~\cite{lin2024fake} & 77.62 / 30.00 & 67.09 / 39.09 & 67.93 / 37.50 & 73.46 / 32.73 & 71.98 / 34.09 & 90.32 / 17.95 &77.47 / 30.00 \\
    RepDFD~\cite{lin2025standing} &83.52 / 24.32 & 61.67 / 41.59 &74.65 / 32.27 &73.62 / 32.27 &62.78 / 40.91 &93.80 / 14.09 & 60.10 / 43.41 \\ 
    Effort~\cite{yan2024effort} & 58.68 / 47.35 & 57.03 / 44.89 & 56.31 / 44.89 & 49.63 / 48.30 & 63.60 / 40.34 & 82.93 / 23.86 & 56.84 / 44.89 \\
    \midrule 
    DiffID~\cite{diff_id} & 75.85 / 32.01 & 78.47 / 28.86 &70.36 / 35.72 &45.26 / 52.97 &64.29 / 39.35 &84.66 / 23.58 &69.51 / 35.56\\
    ICT-Ref~\cite{ict_ref} &63.15 / 39.84 & 70.27 / 32.84 & 72.36 / 31.93 &58.59 / 40.93 &65.36 / 35.73 &74.88 / 24.41 &50.05 / 45.98 \\
    \midrule
    \rowcolor{aliceblue}
    VIPGuard & \textbf{99.69} / \textbf{0.45} & \textbf{98.05} / \textbf{4.80} & \textbf{98.96} / \textbf{1.90} &\textbf{89.03} / \textbf{16.14} &\textbf{97.04} / \textbf{5.36} &\textbf{99.76} / \textbf{0.23} &\textbf{99.27} / \textbf{1.25} \\
    \bottomrule
  \end{tabular}
\end{table}

\paragraph{Comparison with Other LLM-based Methods on VIPEval}
This experiment evaluated the detection capabilities of Multimodal Large Language Models (MLLMs) on VIPEval by comparing various MLLM-based methods, including FFAA~\cite{huang2024ffaa}, an MLLM specialized for face forgery detection, alongside naive MLLMs. Detection performance was measured using Accuracy (ACC) due to the API's binary (Real/Fake) output. As presented in Table ~\ref{tab:fs_acc}, our method consistently outperforms other approaches across all forged image types, demonstrating the effectiveness of identity-specific semantic detection.

\begin{table}[h]
    \centering
    \scriptsize
    \caption{Comparison (ACC (\%)) of our method and other MLLMs on the VIPEval.}
    \label{tab:fs_acc}
    \setlength{\tabcolsep}{7pt}
    \begin{tabular}{l|ccccccccc}
      \toprule
      \textbf{Method} & & &BlendFace  &HifiFace &MobileSwap &UniFace &{ConsistentID} &Arc2Face \\
      \midrule
      
      GPT-4o-2024-08-06\cite{gpt4o} & & &71.85  &84.36 & 94.35 &70.72 &46.70 &50.90 \\
      Gemini-2.5-pro-exp-03-25~\cite{Gemini} & & & 80.87  &82.68  & 91.75 & 83.58 & 69.85 & 79.98 \\
      
      Qwen2.5VL 7B~\cite{bai2025qwen2} & & &50.05 &49.86 &49.99 &50.02 &49.51 &49.69 \\
      LLaMA3.2Vision 11B~\cite{grattafiori2024llama} & & & 44.76 &43.93 &45.71 &54.09 &38.56 &43.61 \\
      FFAA~\cite{huang2024ffaa} & & &88.60 &89.95 &91.36 &89.30 &63.53 &59.89 \\
      \midrule
      \rowcolor{aliceblue}
      VIPGuard & & &\textbf{95.51} &\textbf{95.82}  &\textbf{96.71} & \textbf{95.88}  &\textbf{95.91} &\textbf{89.71}  \\
      \bottomrule
    \end{tabular}
  \end{table}

\vspace{-2mm}
\paragraph{One-shot Performance on other Deepfake detection datasets}
This experiment evaluates our method against several existing ID-aware approaches on established benchmark datasets. The evaluation specifically employed the challenging CelebDF~\cite{li2019celeb} dataset and a CelebDF-related subset of DF40~\cite{yan2024df40}. Owing to limited real sample diversity in these datasets, our method was evaluated in a one-shot setting, using a single real image per user. To mitigate similarity from shared video sources, reference and test images were sampled from different videos. As shown in Table~\ref{tab:oneshot}, our method still achieves competitive performance compared to other approaches. Notably, due to limited real image availability, these results were obtained using VIP-Guard at Stage 2, omitting Stage 3. The results also verify the effectiveness of Stage 2 of our method.

\begin{table}[h]
    \centering
    \tiny
    \caption{Evaluation of frame-level performance (AUC (\%) / EER (\%)) in CelebDF~\cite{li2019celeb} and DF40~\cite{yan2024df40} under one-shot setting. For each identity, only a single real image is available. }
    \begin{tabular}{l|c|ccccccc}
    \toprule
    \multirow{2}{*}{\scriptsize\textbf{Method}} & \multirow{2}{*}{\scriptsize{CelebDF}} & \multicolumn{7}{c}{\scriptsize{DF40}} \\
    \cmidrule(lr){3-9}
     & & BlendFace & E4S & FaceDancer & FSGAN & SimSwap & UniFace & InSwap \\
    \midrule
    Diff-ID &86.66 / 21.79 &76.83 / 28.99 &84.79 / 24.07 &81.04 / 27.02 &82.88 / 23.77 &64.87 / 38.28 &84.04 / 22.53 &59.62 / 43.19 \\
    ICT-Ref &81.86 / 26.18 &76.47 / 30.63 &83.33 / 24.93 &91.57 / 16.38 &73.62 / 32.95 &82.41 / 25.62 &80.34 / 27.07 &69.91 / 35.98 \\
    \midrule
    \rowcolor{aliceblue}
     VIPGuard &87.96 / 20.84 &81.87 / 26.23 &89.74 / 19.63 &86.12 / 23.61 &86.44 / 20.87 &83.65 / 24.92 &86.70 / 22.57 &70.97 / 34.15\\
    \bottomrule
    \end{tabular}
    \label{tab:oneshot}
\vspace{-2mm}
\end{table}

% \subsection{Capability to Recognize Facial Attirbutes}
% \paragraph{Evaluation on the Face Attributes Dataset}
\vspace{-2mm}

\begin{figure}[h]
    \centering
    \includegraphics[width=0.98\linewidth]{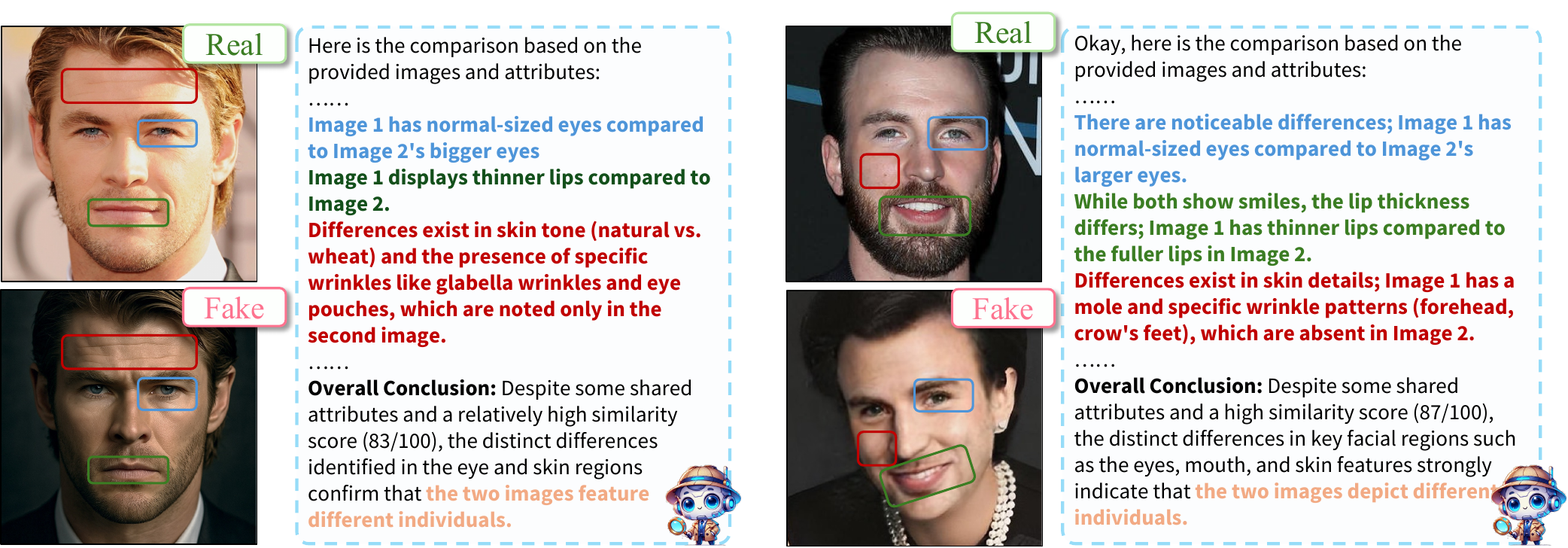}
    \caption{Visual illustration of the analysis of VIP-Guard detecting anomalous local facial attributes for EFS (left) and FS (right). The two real images are sourced from LAION-Face~\citep{laion_face}, while the fake images in the left and right subfigures were generated by GPT-4o~\citep{gpt4o} and HifiFace~\citep{wang2021hififace}, respectively.}
    \label{fig:exp_instance}
    \vspace{-2mm}
\end{figure}

\paragraph{Visual Examples of Our Model Explanations}
Figure~\ref{fig:exp_instance} presents two representative examples illustrating VIPGuard's ability to detect anomalous local facial attributes in deepfake images generated by entire face synthesis (EFS, left) and face swapping (FS, right) techniques. 
Despite the high global facial similarity between the real and fake images, VIPGuard accurately identifies subtle but critical differences in localized regions such as the eyes, lips, and skin textures. For instance, discrepancies in eye size, lip thickness, and specific skin features (e.g., glabella wrinkles, eye pouches, moles, and crow’s feet) are effectively captured by our model. 
These examples highlight two key strengths of VIPGuard: (1) its ability to detect forgery-induced anomalies in local facial regions where current generation methods often fail, and (2) its capacity to reason over these attributes for reliable identity verification, even when global facial appearance is highly similar.

% \subsection{Ablation Study}
% \begin{wraptable}{r}{0.5\textwidth}
% \begin{table}
%     \centering
%     \tiny
%     \caption{The ablation study.}
%     \begin{tabular}{l|cccc}
%     \toprule
%      \textbf{Method} & BlendFace & InSwap & Arc2Face &PuLID \\
%      \midrule
%      Qwen-2.5VL-7B & x  \\
%      \midrule
%      + Stage 3 & \\
%      + Stage 1, 3 & \\
%      + Stage 1, 2 \\
%      + Stage 1, 2, 3 \\
%      \bottomrule
%     \end{tabular}
%     \label{tab:my_label}
% % \end{wraptable}
% \end{table}
\vspace{-2mm}

% \paragraph{Token Number}
% In this experiment, we explore the number of training samples and the VIP token per identity.

\subsection{Effectiveness in Annotation-Free Scenarios}
To validate VIPGuard's practicality, we tested its Stage 3 performance when trained using only VIP images, foregoing textual annotations. 
As shown in Table~\ref{tab:img_only}, the model’s performance remains strong, with the average AUC decreasing only slightly from 98.98\% to 95.86\%. This robustness stems directly from the design of our framework. Stage 2 is responsible for internalizing a vast repository of discriminative knowledge by learning to distinguish between arbitrary image pairs with the aid of textual supervision. 
This foundational pre-training is so effective that the User-specific fine-tuning in Stage 3 requires only visual information to achieve a high degree of accuracy. 
Consequently, this experiment validates VIPGuard as a highly practical and readily deployable framework for real-world applications, including annotation-free scenarios.
The model's ability to achieve such high efficacy under these constraints is a direct testament to the comprehensive and foundational knowledge provided by Stage 2.

\begin{table}[h]
  \centering
  \scriptsize
  \caption{Performance (AUC (\%)) of VIPGuard under different training configurations in Stage 3. \textit{Images + Annotation} denotes training with both images and textual descriptions, while \textit{Only Images} uses visual inputs only.}
  \label{tab:img_only}
  \setlength{\tabcolsep}{17pt}
  \begin{tabular}{l|ccccc}
    \toprule
    \textbf{Variant} & \textbf{BlendFace} & \textbf{InSwap} & \textbf{Arc2Face} & \textbf{PuLID} & \textbf{Average} \\
    \midrule
    \textit{Only Images} & 98.45 & 92.91 & 94.35 & 97.72 & 95.86 \\
    \textit{Images + Annotation} & \textbf{99.48} & \textbf{99.43} & \textbf{98.05} & \textbf{98.96} & \textbf{98.98} \\
    \bottomrule
  \end{tabular}

\end{table}

% \paragraph{Impact of different compositions}
\vspace{-1mm}
\section{Conclusion}
\vspace{-2mm}
This paper proposes VIPGuard, addressing a critical gap in deepfake detection by leveraging known facial identities to enable personalized, accurate, and explainable detection. Unlike traditional detectors that mainly rely on low-level visual artifacts or general-purpose MLLMs lacking identity awareness, VIPGuard integrates fine-grained attribute learning, identity-level discriminative training, and user-specific customization through a unified multimodal framework. 
Combined with our newly proposed VIPBench, which enables rigorous evaluation across diverse and advanced forgery types, our approach demonstrates clear superiority in both detection performance and explainability, offering a robust and scalable solution for safeguarding high-risk individuals against identity-based deepfakes.

\vspace{-2mm}
\paragraph{Content Structure of the Appendix.}
Due to page constraints, we include additional analyses and experiments in the Appendix, containing comprehensive ablation studies (Appendix ~\ref{app:ablation}), robustness evaluation (Appendix ~\ref{app:robust}), adaptive choice of VIP Token (Appendix ~\ref{app:ada_vip}), more visual examples of model explanations (Appendix ~\ref{app:Visualization}), details of dataset construction (Appendix ~\ref{app:data}), an ethical statement (Appendix ~\ref{app:ethics}), and limitations and future work (Appendix ~\ref{app:limitation}).
\textit{\textbf{For further details, please refer to the Appendix.}}

\clearpage

\begin{ack}
Dr Bin Li was supported in part by NSFC and in part by Shenzhen R\&D Program (Grant JCYJ20250604181211016, SYSPG20241211174032004). 
Dr Weixiang Li was supported in part by NSFC (Grant 62202310, 62572328) and in part by Guangdong Basic and Applied Basic Research Foundation (Grant 2025A1515010292).
\end{ack}
% Dr Bin Li was supported in part by NSFC (Grant U23B2022, U22B2047) and in part by Guangdong Provincial Key Laboratory (Grant 2023B1212060076). 
% Dr Weixiang Li was supported in part by NSFC (Grant 62202310, 62572328) and in part by Guangdong Basic and Applied Basic Research Foundation (Grant 2025A1515010292).
% This work was supported in part by NSFC (Grant U23B2022, U22B2047, 62202310, 62572328); in part by Shenzhen R\&D Program (Grant JCYJ20250604181211016, SYSPG20241211174032004); in part by Guangdong Basic and Applied Basic Research Foundation (Grant 2025A1515010292).

% \section{Ethics Statement}
% This study involves facial forgery and may raise privacy concerns.
% We affirm that the source images in all proposed datasets are derived from publicly available open-source datasets~\cite{laion_face, faceid6m, crossfaceid}.
% Our research is dedicated to protecting facial privacy and is intended solely for ethical academic use.

\bibliographystyle{plain}  % 或 plain, unsrt, acm, apalike, etc.
\bibliography{ref}

\clearpage
\appendix
\newpage

\part*{Appendix}
\addcontentsline{toc}{part}{Appendix}
\label{sec:appendix}
\section{Ethics Statement}
\label{app:ethics}
% This study involves facial forgery and may raise privacy concerns.
% We affirm that the source images in all proposed datasets are derived from publicly available open-source datasets~\cite{laion_face, faceid6m, crossfaceid}.
% Our research is dedicated to protecting facial privacy and is intended solely for ethical academic use.
This study involves facial forgery detection and may raise privacy and ethical concerns related to the use of human facial data. 
We affirm that all source images used in our datasets are obtained from publicly available and legally compliant open-source datasets~\cite{laion_face, faceid6m, crossfaceid}, which are intended for academic research purposes. 
Our research is dedicated to advancing facial privacy protection and deepfake detection. The developed models and datasets are intended solely for ethical, academic use and will not be released for commercial or malicious purposes.
This study follows the ethical guidelines provided by NeurIPS Code of Ethics, and does not involve any personally identifiable information collected directly by the authors.
Our VIPBench dataset is released under the Creative Commons Attribution-NonCommercial (CC BY-NC) license (more details can be seen in \url{https://creativecommons.org/licenses/by-nc/4.0/}). Furthermore, access to the dataset will be managed through a request form (hosted on HuggingFace) to monitor and control its usage. All interested parties are required to complete the form, and each request will be manually reviewed to help prevent potential misuse.

\section{Boarder Impact}
\label{app:impact}
The research presented in this paper introduces VIPGuard, a multimodal framework for personalized deepfake detection, and VIPBench, a benchmark dataset for evaluating identity-aware detectors. This work addresses a critical and growing societal challenge: the malicious misuse of AI-generated media, particularly in the form of deepfakes targeting specific individuals such as celebrities.
Our method promotes positive societal impact in several key areas:
\begin{itemize}
    \item \textbf{Mitigating Disinformation and Identity Theft:} Deepfakes pose a significant threat by enabling the creation of highly realistic fake media, which can be weaponized to harm reputations, manipulate public perception, or conduct fraud. VIPGuard offers a targeted defense mechanism, enhancing the security and privacy of individuals by providing personalized and explainable detection tools.
    \item \textbf{Promoting Safer AI Ecosystems:} By releasing VIPBench, we aim to catalyze further research in robust, personalized detection. Public benchmarks encourage accountability, replication, and the development of defenses that are grounded in real-world threats. Furthermore, we will implement a review-based distribution process for the dataset to ensure its lawful and responsible use.
\end{itemize}

\section{Limitation and Future Works}
\label{app:limitation}
VIPGuard is designed to protect the identities of target users against deepfakes. While the method demonstrates strong generalization capabilities, it does not yet fully exploit additional modalities such as audio, 3D facial models, temporal consistency, and other complementary cues. For instance, these modalities could significantly improve the detection of audio-driven manipulation techniques, which rely heavily on temporal dynamics and inter-frame consistency. Beyond facial images, incorporating such prior information could further enhance detection performance. As part of future work, we plan to integrate these complementary modalities to strengthen VIPGuard’s effectiveness.
In addition, we recognize two key limitations observed during evaluation: (1) \textbf{Significant variations} in head pitch and yaw angles, such as noticeable rotations exceeding $50^\circ$, can affect facial features and identity information. However, this limitation is shared by most face-based systems, including standard face recognition models. (2) \textbf{Significant age differences} between training and test data can lead to discrepancies in performance. For instance, training may rely on younger facial images while testing may involve older ones. This issue, arising from identity shifts across age, can be alleviated by expanding and diversifying the dataset. Inspired by these observations, we plan to address these potential failure cases in future work by incorporating 3D facial information, learning identity-related temporal cues from videos, and exploring other complementary sources of information.

\section{Experiments}

\subsection{Implementation Details}
\label{app:details}
This paper adopted the pre-trained Qwen-2.5-VL-7B model~\cite{bai2025qwen2} as the backbone, adhering to its default pre-processing settings. 
Input images were resized to $448 \times 448$ when larger than this size.
The model was optimized using the Adam optimizer with a cosine learning rate decay schedule, starting from an initial learning rate of 3e-5.
To accommodate GPU memory limitations, the equivalent batch size was maintained at 72 for both Stage 1 and Stage 2 by applying gradient accumulation.
In Stage 3, the effective batch size was reduced to 8 and the initial learning rate was set to 1.
All training was performed using mixed-precision computation within the open-source Swift\footnote{https://github.com/modelscope/ms-swift} framework.
The model was trained for 2 epochs in Stage 1, and for 1 epoch each in Stage 2 and Stage 3.
For the evaluation in the main paper, we obtained the AUC and EER metrics by computing the normalized probability of output logits in MLLM's prediction head for the words `\texttt{Yes}' and `\texttt{No}'. 
During evaluation, VIPGuard was only required to output the final prediction (i.e., `\texttt{Yes}' or `\texttt{No}') without providing any explanatory content.

\subsection{Ablation Study}
\label{app:ablation}
In this section, we conduct a series of ablation studies to comprehensively evaluate the effectiveness of different components and design choices in our proposed VIPGuard framework. 
The AUC ($\%$) metric was selected to exhibit the performance in detecting multiple facial forgeries.

\begin{table}
  \footnotesize
  \centering
  \caption{An ablation study on the effectiveness (AUC ($\%$)) of different components in VIPGuard }
  \label{tab:composition}
  \setlength{\tabcolsep}{7pt}
  \begin{tabular}{lp{4.5cm}|ccccc}
    \toprule
    \textbf{Variant} &\textbf{Description} & \textbf{BlendFace} & \textbf{InSwap} & \textbf{Arc2Face} &\textbf{PuLID} \\
    \midrule
    Baseline &Qwen-2.5-VL-7B & 51.00 & 49.79 & 49.85 & 50.30 \\
    \midrule
    + Stage 3 &{\small Only performing User-Specific Customization} &71.96 &59.10 &62.12 &68.18 \\
    \\
    + Stage 1, 3 &{\small Enhancing facial understanding and then performing User-Specific Customization} &96.50 &87.38 &88.95 &94.11 \\
    \\
    + Stage 1, 2 &{\small Enhancing facial understanding and general identity discrimination between arbitrary face pairs} &90.82 &73.44 &77.25 &71.08 \\
    \\
    + Stage 1, 2, 3 &{\small Enhancing facial understanding, general identity discrimination, and User-Specific Customization}  &99.48 &96.40 &98.05 &98.96 \\
    \bottomrule
  \end{tabular}
\end{table}

\paragraph{Exploration for the Composition in VIPGuard}
\label{app:components}
In this paper, we propose a three-stage learning framework VIPGuard to respectively improve MLLM's face understanding, fine-grained discrimination between arbitrary face pairs, and fine-grained discrimination for VIPs.
To evaluate the contribution of each stage, we conduct an ablation study using different combinations of the three stages, as demonstrated in Table ~\ref{tab:composition}. The base model, Qwen-2.5-VL-7B, demonstrates near-random performance across all datasets, indicating its limited native capability in detecting forged faces.
Adding only Stage 3, which introduces user-specific tuning, moderately improves performance (e.g., +20.96 on BlendFace and +17.88 on PuLID), but results remain suboptimal due to the model’s lack of foundational face understanding. Introducing Stage 1 alongside Stage 3 yields substantial gains across all benchmarks, confirming that basic face comprehension is essential for effective user-specific customization.
Combining Stages 1 and 2, without user-specific tuning, enhances general discrimination (e.g., +45.50 on BlendFace compared to the base model), but is still inferior to the full configuration. Notably, the detection for VIP users was conducted under a one-shot setting, where the MLLM compares a single reference image of the target user with the queried image.  Without user-specific customization, the MLLM cannot accurately capture the nuanced identity traits of the VIP user, resulting in limited detection performance.
The complete VIPGuard pipeline (Stages 1, 2, and 3) achieves the highest performance across all datasets, with detection rates exceeding 96$\%$ in most cases (e.g., 99.48 on BlendFace, 98.96 on PuLID), highlighting the critical synergy among the three stages.
These results demonstrate that VIPGuard’s effectiveness hinges on the sequential integration of facial understanding, general identity discrimination, and user-specific adaptation. Each component contributes uniquely, and omitting any stage leads to measurable performance degradation.

\begin{figure}
    \centering
    \includegraphics[width=0.8\linewidth]{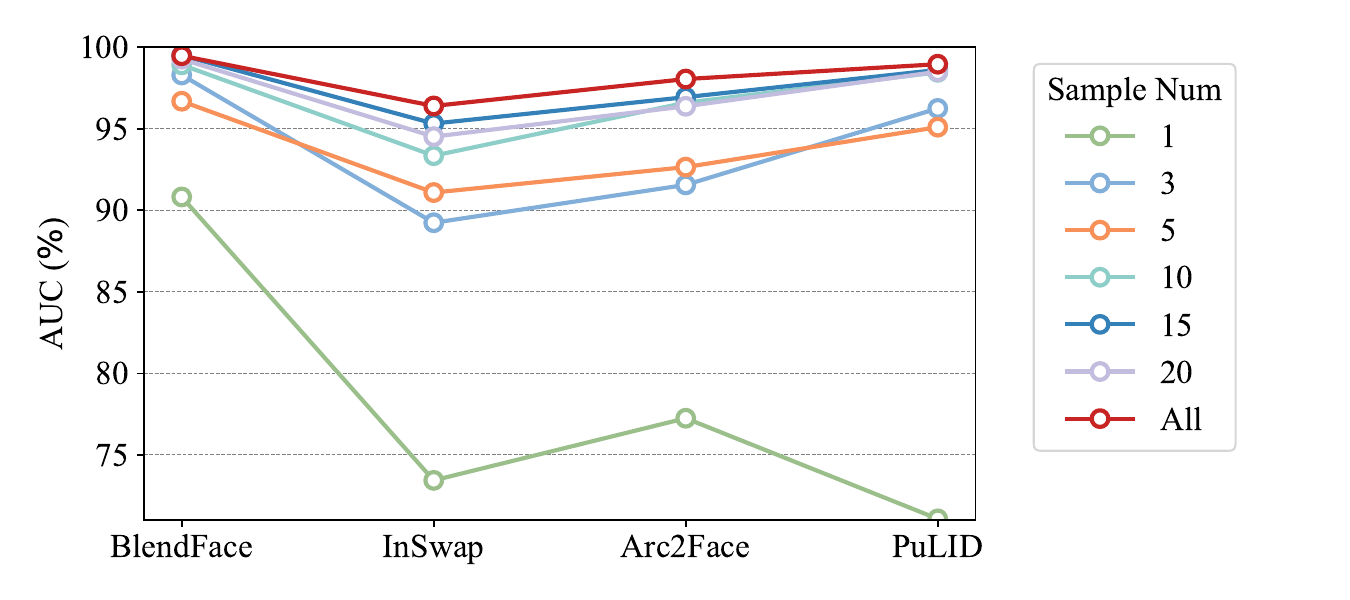}
    \caption{Ablation study on VIPGuard performance (AUC ($\%$)) with varying numbers of available authentic images of target users. `\texttt{All}' refers to using all available authentic images, typically ranging from 20 to 40 images for each identity in VIPBench.}
    \label{fig:sample_num}
\end{figure}

\paragraph{Impact of the Number of Available Authentic Images for Target Users (VIPs)}
\label{app:vip_num}
While abundant negative samples can be obtained from real images of different identities, the number of accessible authentic images for a target user limits the positive samples available for Stage~3 training, thereby directly impacting the model's user-specific detection capability.
In this experiment, we varied the quantity of available authentic images for a specific user and explored its effect on VIPGuard's performance (Figure ~\ref{fig:sample_num}).
When only one authentic image was accessible, the scarcity of positive samples precluded Stage~3 training; consequently, the Stage~2 trained model was used directly for detection without User-Specific Customization.
The results indicate that VIPGuard's detection capability improves as more authentic images of the specific user become available.
Compared to directly using the Stage~2 model, employing Stage~3 with even just three authentic images (positive samples) yielded a substantial improvement in detection performance.
Furthermore, as the number of accessible user samples increased, detection performance across various forgery types showed consistent improvement.
% \begin{table}
%   \centering
%   \caption{The performance of VIPGuard by varing the available authentic images of target users.}
%   \label{tab:sample_num}
%   \begin{tabular}{lccccccc}
%     \toprule
%     \textbf{Sample Num} & BlendFace & InSwap & Arc2Face &PuLID  \\
%     \midrule
%     1(One Reference) &90.82 &73.44 &77.25 &71.08  \\
%     3 & 98.30 & 89.24 &91.56 &96.25 & \\
%     5 & 96.69 &91.09 &92.65 &95.10 \\
%     10 &98.92 &93.35 &96.60 &98.46  \\
%     15 &99.50 &95.32 &96.94 &98.61 \\
%     20 &99.27 &94.52 &96.38 &98.50 \\
%     All &99.48 &96.40 &98.05 &98.96 \\

%     \bottomrule
%   \end{tabular}
% \end{table}

\begin{figure}
    \centering
    \includegraphics[width=0.8\linewidth]{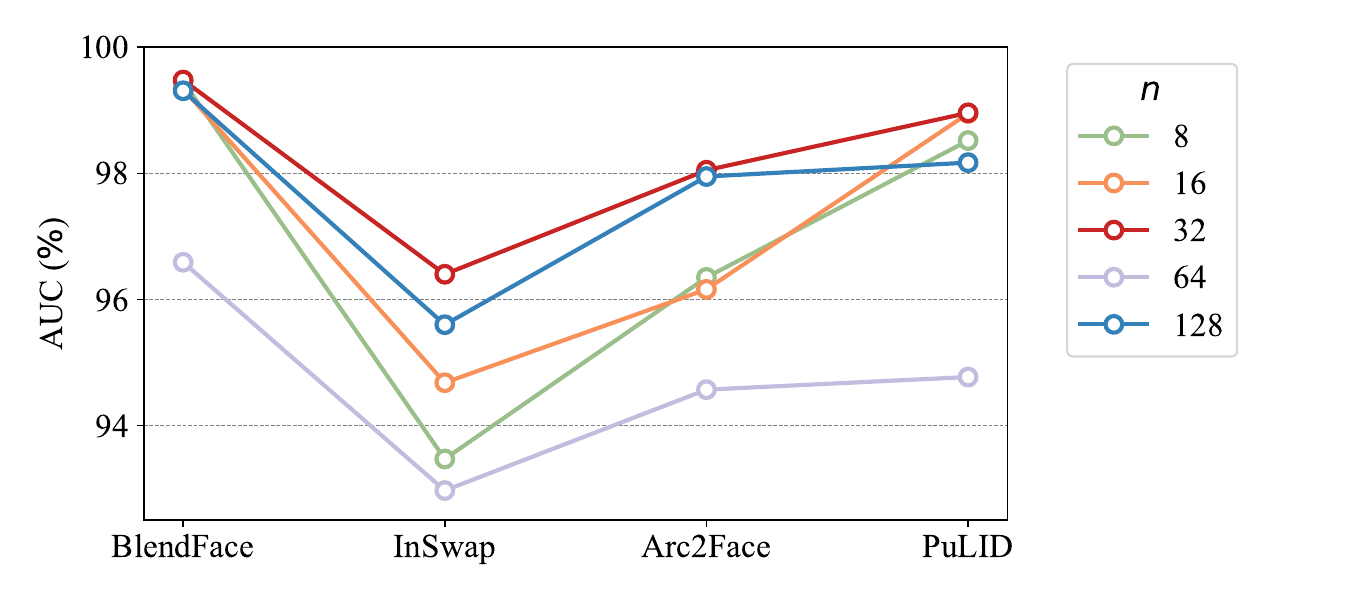}
    \caption{Ablation study on VIPGuard performance (AUC ($\%$)) with varying the size $n$ of the VIP token. }
    \label{fig:token_num}
\end{figure}

\paragraph{Impact of the Size of VIP tokens in Stage3}
\label{app:token_num}
In Stage~3, the VIP Token $\mu$ is of size $n \times d$, where $n$ denotes the number of VIP Tokens and $d$ represents the feature dimensionality.
We investigated the impact of varying $n$ on performance.
As shown in Figure ~\ref{fig:token_num}, increasing $n$ up to 32 improved performance, reaching an optimum at $n=32$.
However, further increasing $n$ to 64 or 128 resulted in a performance decline, potentially due to overfitting on the limited training samples. Therefore, we finally set $n$ to $32$.
% \begin{table}
%   \centering
%   \caption{The performance of VIPGuard by varing the sizes of the VIP token.}
%   \label{tab:token_num}
  
%   \begin{tabular}{lccccccc}
%     \toprule
%     $n$ & BlendFace & InSwap & Arc2Face &PuLID \\
%     \midrule
%     8 &99.46 &93.47 &96.35 &98.52  \\
%     16 &99.34 &94.68 &96.16 &98.95 \\
%     32 &99.48 &96.40 &98.05 &98.96 \\
%     64 &96.59 &92.97 &94.57 &94.77 \\
%     128 &99.31 &95.60 &97.95 &98.17 \\
%     \bottomrule
%   \end{tabular}
% \end{table}

\paragraph{Performance with Different Backbones}
\label{app:backbone}
To further validate the universality of VIPGuard, we evaluated its performance using different multimodal large language model (MLLM) backbones.
As shown in Table~\ref{tab:backbone}, we replaced the backbone with LLaMA-3.2-Vision-11B~\citep{llama32}, Qwen-2.5-VL-3B~\citep{bai2025qwen2}, and Qwen-2.5-VL-7B~\citep{bai2025qwen2}, respectively.
Across all configurations, VIPGuard consistently achieved strong and stable performance improvements over the baseline model, demonstrating its adaptability to diverse MLLM architectures and its great detection performance across different generators.

\begin{table}
  \centering
  \footnotesize
  \caption{Performance Comparison (AUC (\%)) of VIPGuard with Different MLLMs.}
  \label{tab:backbone}
  \setlength{\tabcolsep}{11pt}
  \begin{tabular}{l|ccccccc}
    \toprule
    \textbf{Method} & \textbf{BlendFace} & \textbf{InSwap} & \textbf{Arc2Face} &\textbf{PuLID} &\textbf{Average}  \\
    \midrule
    \rowcolor{gray!10}
    \multicolumn{6}{l}{\textbf{\textit{Baseline}}} \\
    \addlinespace[3pt]
    Qwen-2.5-VL-7B (Baseline) &51.00 &49.79 &49.85 &50.30 &50.24  \\
    \midrule
    \rowcolor{gray!10}
    \multicolumn{6}{l}{\textbf{\textit{VIPGuard}}} \\
    \addlinespace[3pt]
    LLaMA-3.2-Vision-11B &96.27 &88.97 &\textbf{99.00} &\textbf{99.52} &95.94 \\
    Qwen-2.5-VL-3B &90.71 &87.85 &87.82 &97.94 &91.08 \\
    Qwen-2.5-VL-7B &\textbf{99.48} &\textbf{96.40} &98.05 &98.96 &\textbf{98.23} \\
    \bottomrule
  \end{tabular}
\end{table}

\paragraph{Impact of the Face Recognition Models}
\label{app:facemodel}
In the proposed method, the face recognition model is utilized to provide the global facial information and output a similarity score.
To determine which face-recognition backbone is most suitable for computing similarity in VIPGuard, we conducted a comparative evaluation of several popular models, including CosFace~\citep{cosface}, ArcFace~\citep{arcface}, and TransFace~\citep{transface}. 
Each model was integrated into the VIPGuard framework to perform identity similarity estimation, and the performance was measured by AUC (\%) across four representative forgery datasets: BlendFace~\citep{blendface}, InSwap~\citep{inswapper}, Arc2Face~\citep{arc2face}, and PuLID~\citep{guo2024pulid}. 
As summarized in Table~\ref{tab:face_model}, TransFace consistently achieved the highest accuracy among the tested models, demonstrating its superior capability in capturing discriminative and fine-grained identity representations. 
Therefore, we adopt TransFace as the default face-recognition model in our experiments.

\begin{table}[h]
  \centering
  \footnotesize
  \caption{Performance (AUC (\%)) of VIPGuard using different face-recognition models as similarity components. TransFace exhibits the best overall performance and is thus used by default in our framework.}
  \label{tab:face_model}
  \setlength{\tabcolsep}{15pt}
  \begin{tabular}{l|ccccc}
    \toprule
    \textbf{Face Model} & \textbf{BlendFace} & \textbf{InSwap} & \textbf{Arc2Face} & \textbf{PuLID} & \textbf{Average} \\
    \midrule
    CosFace & 95.12 & 83.32 & 83.31 & 91.74 & 88.37 \\
    ArcFace & 98.14 & 83.17 & 94.41 & 97.88 & 93.40 \\
    TransFace & \textbf{99.48} & \textbf{96.40} & \textbf{98.05} & \textbf{98.96} & \textbf{98.23} \\
    \bottomrule
  \end{tabular}
\end{table}

\subsection{Adaptive VIP Token Selection}
\label{app:ada_vip}
In practical scenarios, it is common to encounter multiple VIP users requiring protection, which poses the challenge of automatically identifying and selecting the appropriate VIP token without manual effort.
To address this, we develop an adaptive variant, Adaptive VIPGuard, that autonomously selects the most relevant VIP token, thereby improving the scalability and usability of the proposed framework in real-world applications.
 Specifically, we first leverage a face recognition model to identify the VIP user whose facial features are most similar to those in the input image. 
Subsequently, Adaptive VIPGuard utilizes the corresponding VIP token to perform identity-specific forgery detection.
As shown in Table ~\ref{tab:adaVIP}, the comparable performance between Adaptive VIPGuard and the original VIPGuard demonstrates the robustness and practicality of our framework in real-world applications.

\begin{table}[h]
  \centering
  \footnotesize
  \caption{Comparison of AUC (\%) for adaptive VIP token selection in VIPEval, where Adaptive VIPGuard refers to the automatic selection of the VIP token without manual intervention.}
  \label{tab:adaVIP}
  \setlength{\tabcolsep}{13pt}
  \begin{tabular}{l|ccccccc}
    \toprule
    \textbf{Method} & \textbf{BlendFace} & \textbf{InSwap} & \textbf{Arc2Face} &\textbf{PuLID} &\textbf{Average}  \\
    \midrule
    VIPGuard &{99.48} &{96.40} &98.05 &98.96 &{98.23} \\
    Adaptive VIPGuard &99.31 &96.14 &97.63 &98.95 &98.01 \\
    \bottomrule
  \end{tabular}
\end{table}

\subsection{Robustness Evaluation}
\label{app:robust}
We conducted experiments under common image degradations, including Gaussian noise, Gaussian blurring, and JPEG compression, to evaluate the robustness of VIPGuard. 
As shown in Table~\ref{tab:robustness}, VIPGuard remains highly effective even under severe degradation levels. 
Because the detector primarily relies on high-level semantic cues such as facial structure and identity-related features, it is inherently less sensitive to low-level pixel distortions.
Table~\ref{tab:robustness} presents the average AUC performance of VIPGuard in detecting BlendFace~\citep{blendface}, InSwap~\cite{inswapper}, Arc2Face~\citep{arc2face}, and PuLID~\cite{guo2024pulid} forgeries under different degradation intensities. 
The results demonstrate minimal performance decline across all settings, confirming the model’s robustness and strong generalization capacity. 
The detailed degradation configurations are provided in Table~\ref{tab:degradation}.
These results confirm that VIPGuard maintains stable detection performance even when subjected to considerable image degradation, demonstrating its robustness and practicality for real-world deployment.

\begin{table}[h]
\centering
\footnotesize
\caption{VIPGuard’s performance (AUC (\%)) under different image degradations. Higher levels indicate stronger degradation. The results represent the average AUC across BlendFace, InSwap, Arc2Face, and PuLID.}
\label{tab:robustness}
\setlength{\tabcolsep}{17pt}
\begin{tabular}{l|ccc}
\toprule
\textbf{Level} & \textbf{Gaussian Noise (Color)} & \textbf{Gaussian Blurring} & \textbf{JPEG Compression} \\
\midrule
None & 98.23 & 98.23 & 98.23 \\
1 & 97.07 & 98.17 & 98.03 \\
2 & 96.53 & 98.16 & 98.10 \\
3 & 94.12 & 98.05 & 97.78 \\
\bottomrule
\end{tabular}
\end{table}

\begin{table}[h]
\centering
\footnotesize
\caption{Degradation configurations for robustness evaluation. Gaussian noise ($\mathcal{N}(0, \sigma^2)$) was applied in YCbCr space; Gaussian blurring was defined by kernel size ($K$) and standard deviation ($\sigma$); JPEG compression was applied using different quality factors (QF).}
\label{tab:degradation}
\setlength{\tabcolsep}{16pt}
\begin{tabular}{l|ccc}
\toprule
\textbf{Level} & \textbf{Gaussian Noise (Color)} & \textbf{Gaussian Blurring} & \textbf{JPEG Compression} \\
\midrule
1 & $\mathcal{N}(0, 8^2)$ & $K=(7,7), \sigma=1$ & QF 90 \\
2 & $\mathcal{N}(0, 11^2)$ & $K=(13,13), \sigma=2$ & QF 60 \\
3 & $\mathcal{N}(0, 18^2)$ & $K=(21,21), \sigma=3$ & QF 30 \\
\bottomrule
\end{tabular}
\end{table}

\subsection{Explanation Examples}
\label{app:Visualization}
\begin{figure}[b]
    \centering
    \includegraphics[width=1\linewidth]{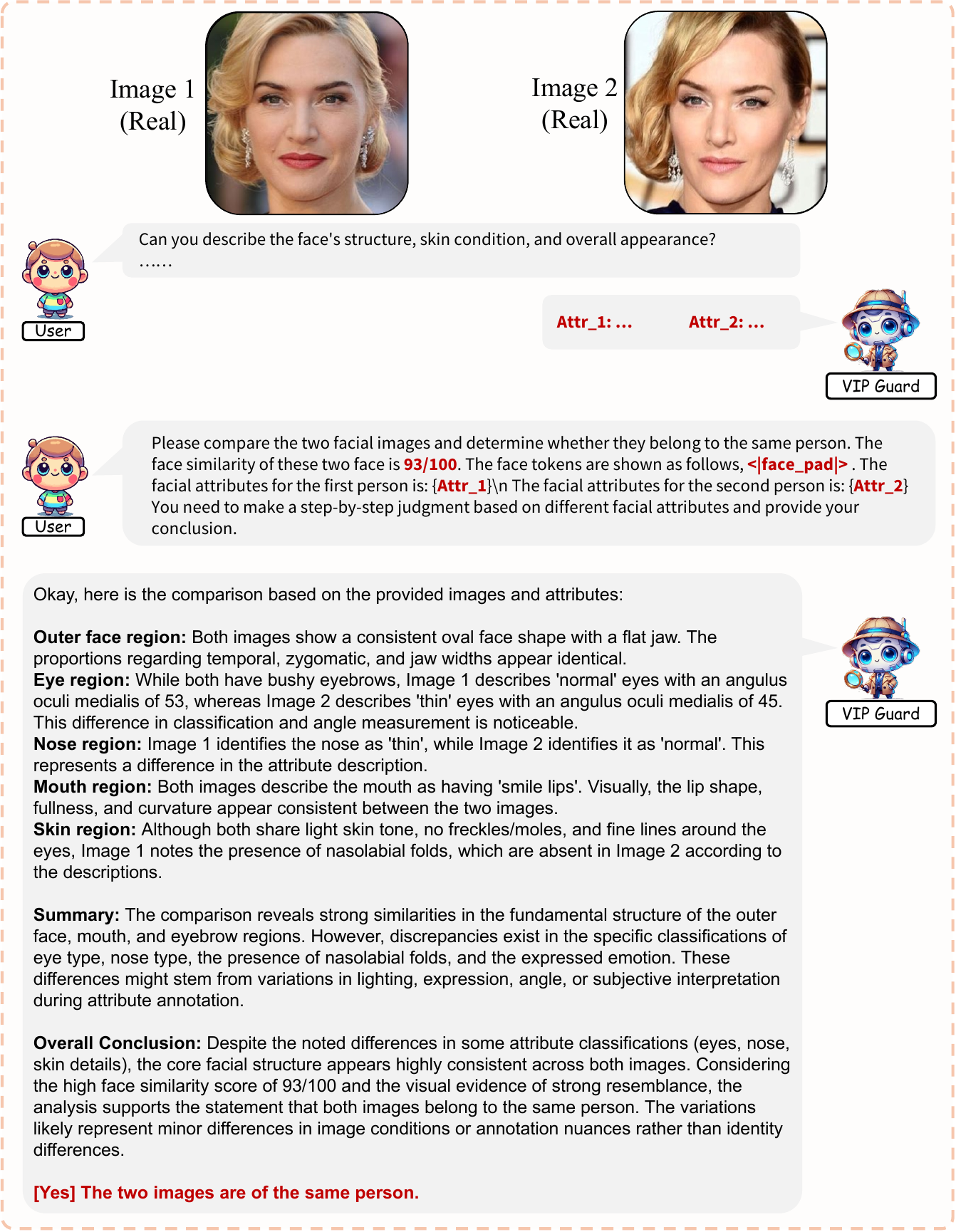}
    \caption{An example of VIPGuard in detecting the real faces. The image 1 and image 2 are from LAION-Face~\citep{laion_face}.}
    \label{fig:exp_real}
\end{figure}

\begin{figure}
    \centering
    \includegraphics[width=1\linewidth]{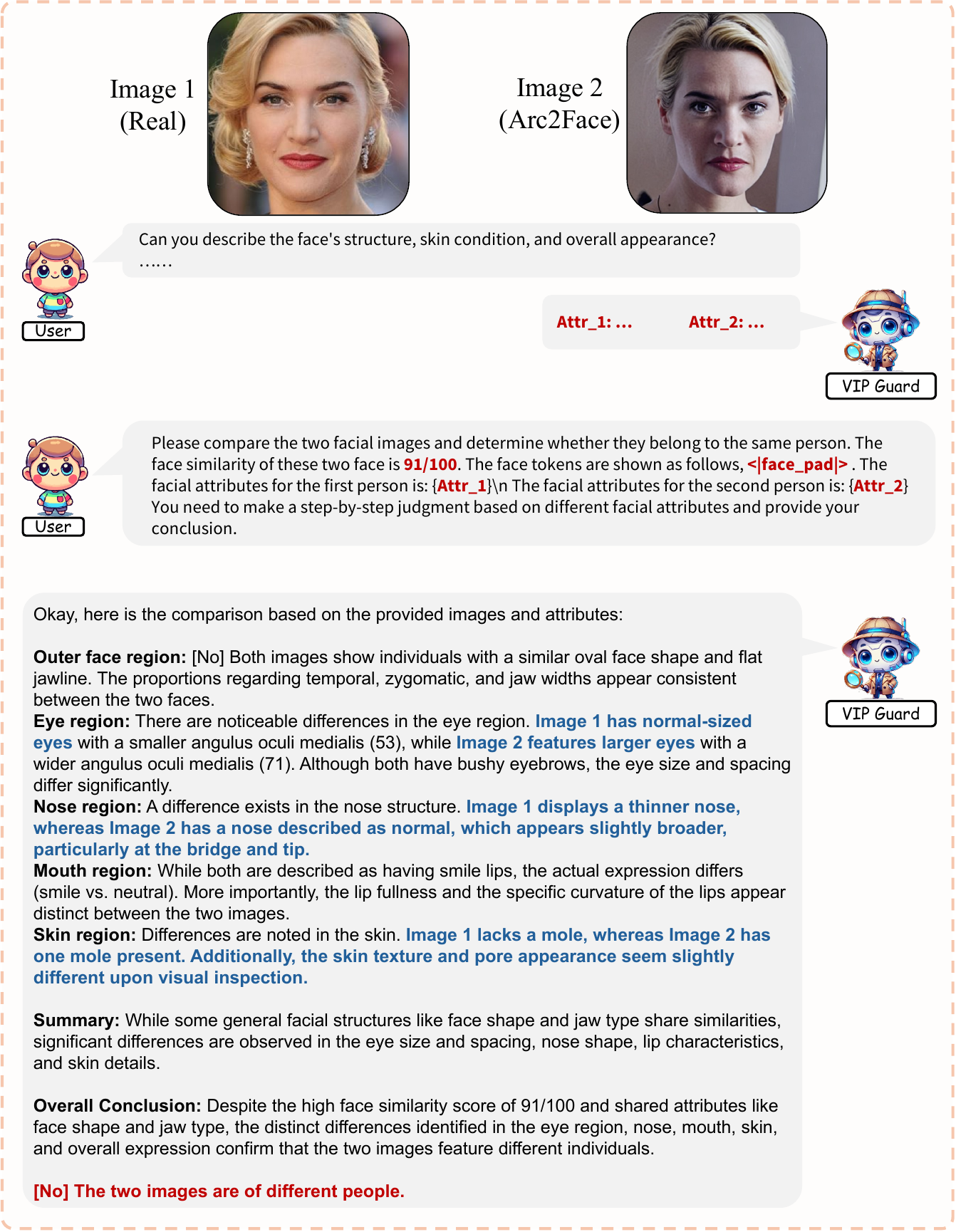}
    \caption{An example of VIPGuard in detecting the forgery generated by Arc2Face~\cite{arc2face}. The image 1 is from LAION-Face~\citep{laion_face} and the image 2 was generated by Arc2Face.}
    \label{fig:exp_1}
\end{figure}

\begin{figure}
    \centering
    \includegraphics[width=1\linewidth]{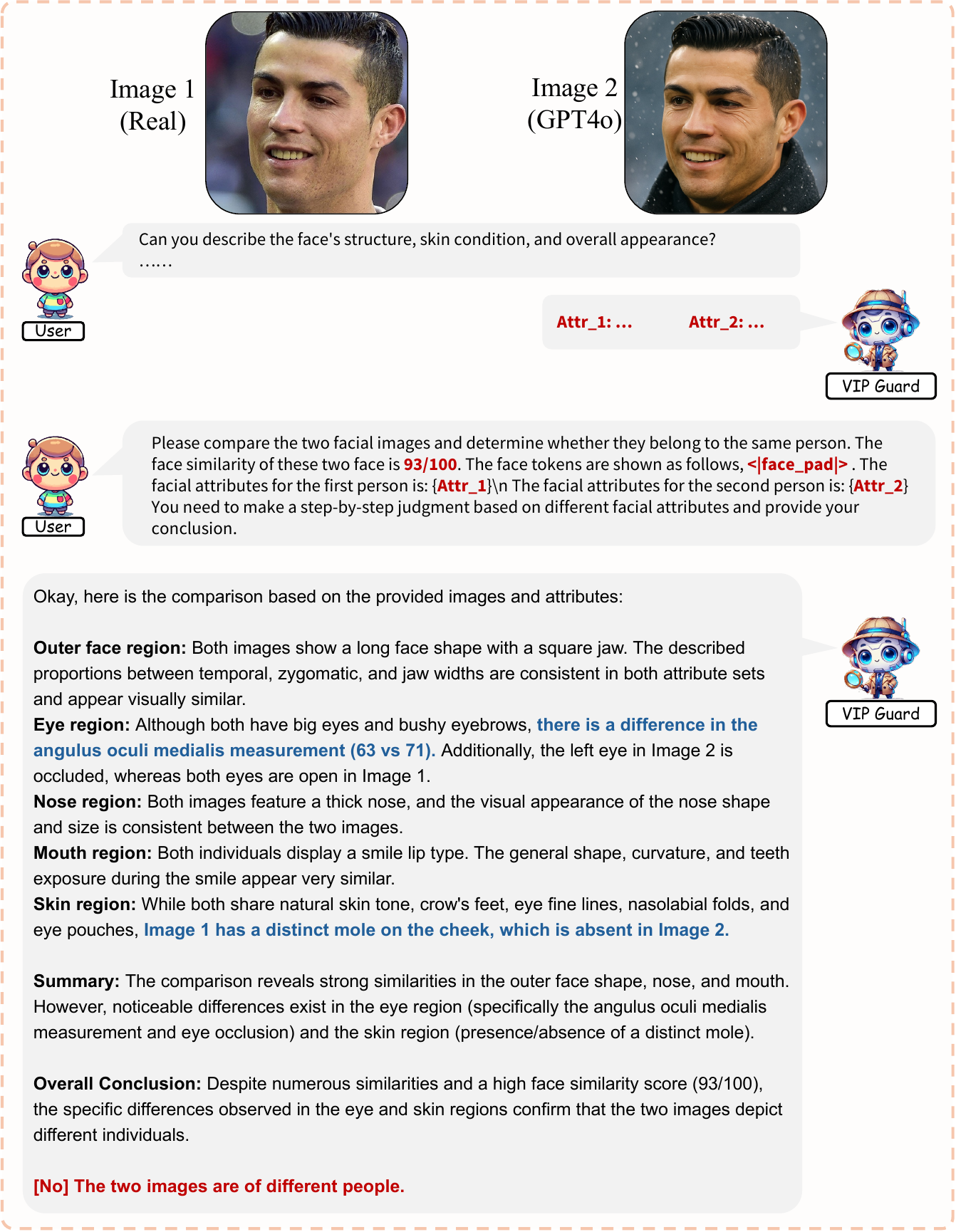}
    \caption{An example of VIPGuard in detecting the forgery generated by GPT-4o~\cite{gpt4o}. The image 1 is from CrossFaceID~\citep{crossfaceid} and the image 2 was generated by GPT-4o.}
    \label{fig:exp_2}
\end{figure}

\begin{figure}
    \centering
    \includegraphics[width=1\linewidth]{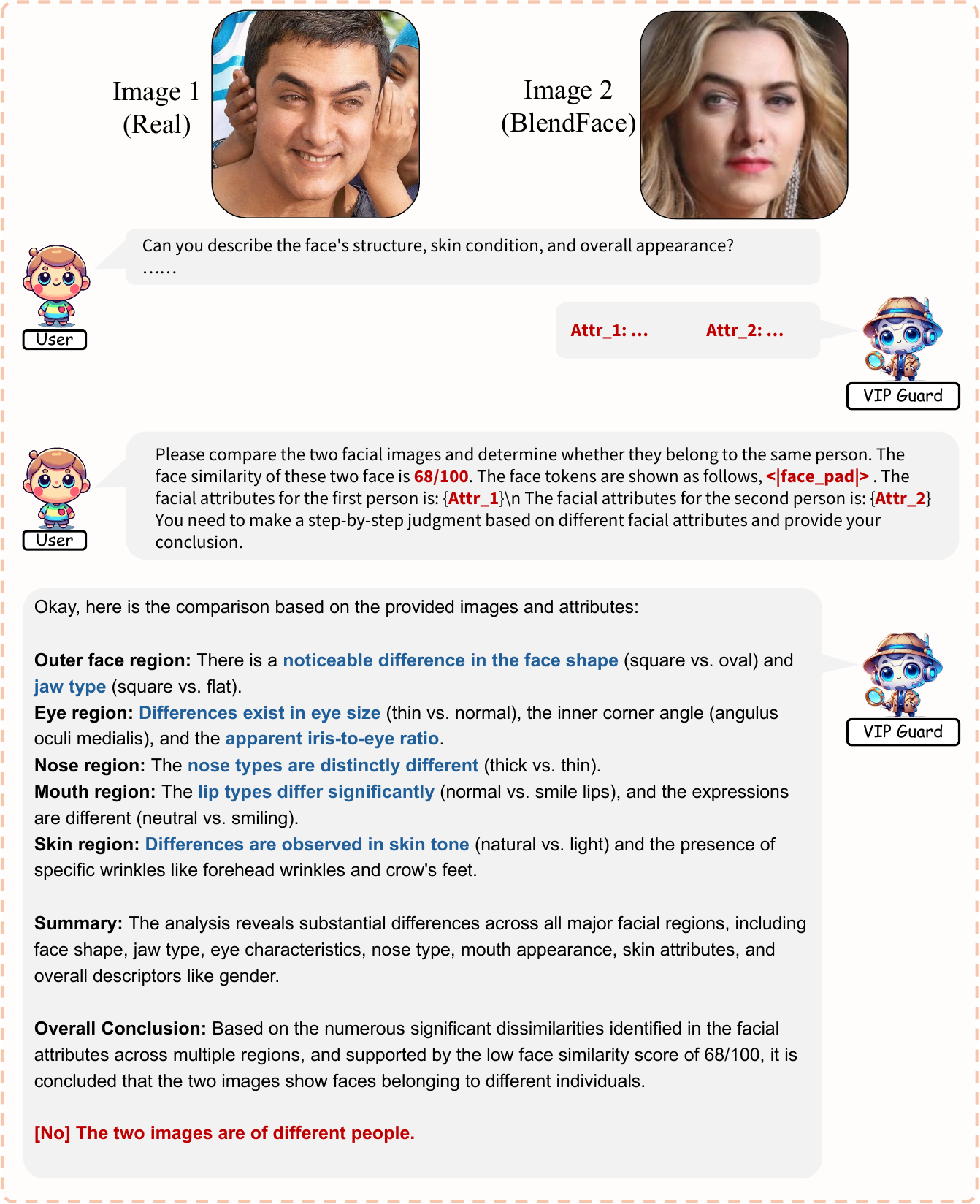}
    \caption{An example of VIPGuard in detecting the forgery generated by BlendFace~\cite{blendface}. The image 1 is from FaceID-6M~\citep{faceid6m} and the image 2 was generated by BlendFace.}
    \label{fig:exp_3}
\end{figure}

We additionally supplement some explanation results, which are exhibited in 
Figures~\ref{fig:exp_real},~\ref{fig:exp_1},~\ref{fig:exp_2}, and~\ref{fig:exp_3}.

\clearpage


\clearpage

\section{Dataset Construction}
\label{app:data}
We introduce a new dataset, VIPBench, designed to provide a more comprehensive evaluation of Identity-aware Deepfake Detection methods~\cite{id1,id2,id3,ict_ref,diff_id}.
In this section, we detailed describe the construction pipeline of VIPBench, which contain three part-Facial Attributes Description Dataset $\mathcal{D}_{FA}$, Identity Discrimination Dataset $\mathcal{D}_{ID}$, and VIPEval $\mathcal{D}_{Eval}$. 
The detailed process encompasses data collection\&preprocessing and the construction of visual question answering (VQA) components, which are shown below.

\subsection{Facial Attributes Description Dataset}
\label{subsec:favqa}
\begin{figure}
    \centering
    \includegraphics[width=1\linewidth]{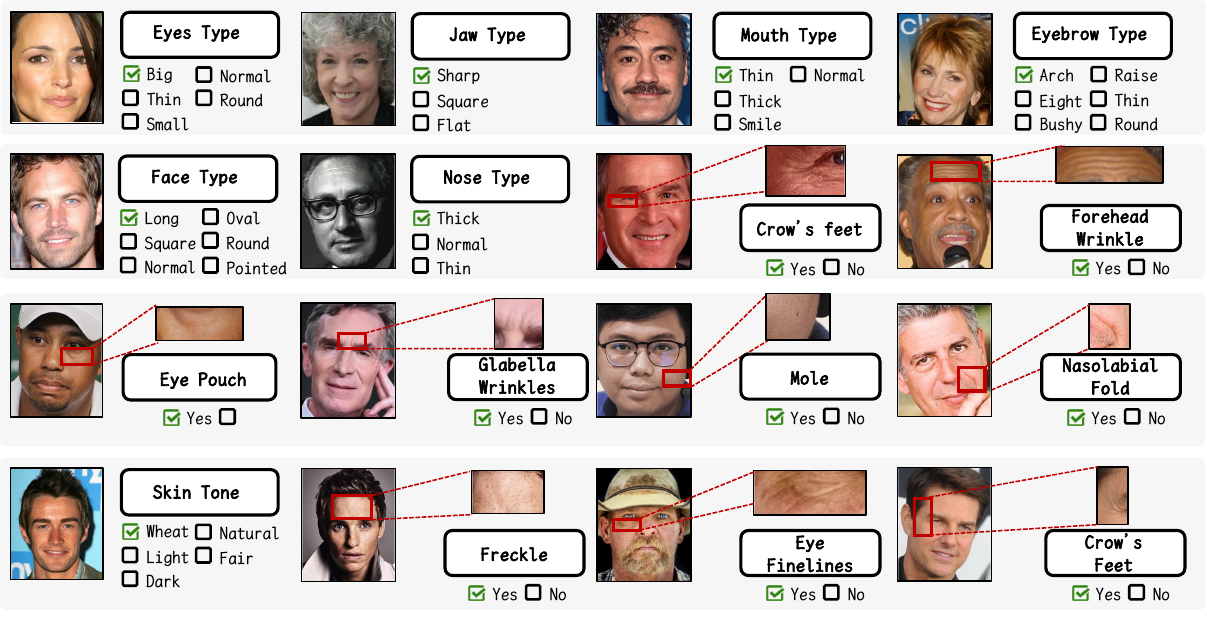}
    \caption{Visualization of different facial attributes. All the images are from LAION-Face~\citep{laion_face}}
    \label{fig:FA}
\end{figure}

\begin{figure}
    \centering
    \includegraphics[width=0.8\linewidth]{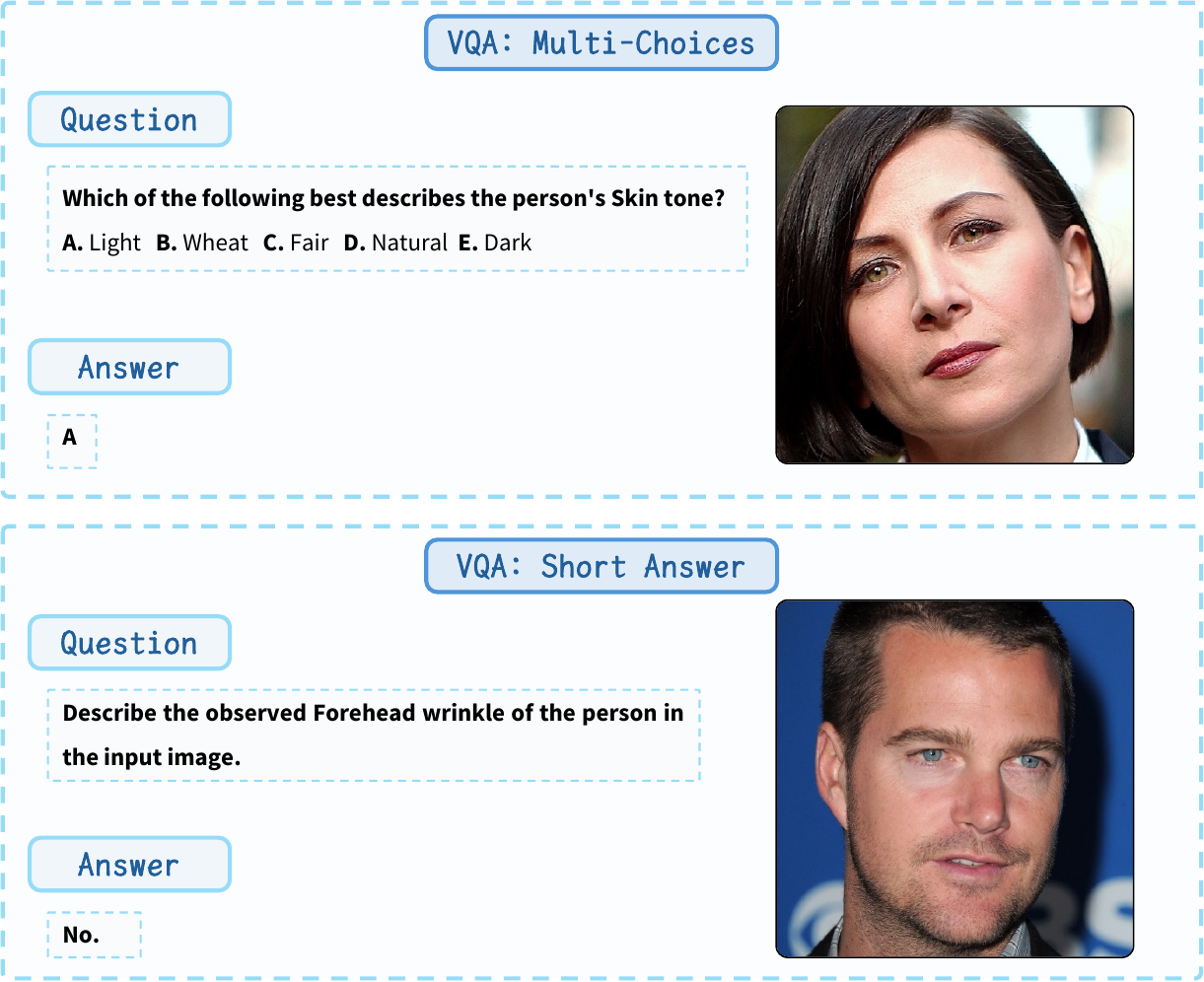}
    \caption{Examples of our VQA data formats, including a multiple-choice question and a short-answer question based on facial attributes. The two images are from LAION-Face~\citep{laion_face}}
    \label{fig:choice}
\end{figure}

\begin{figure}
  \centering
  \includegraphics[width=1\linewidth]{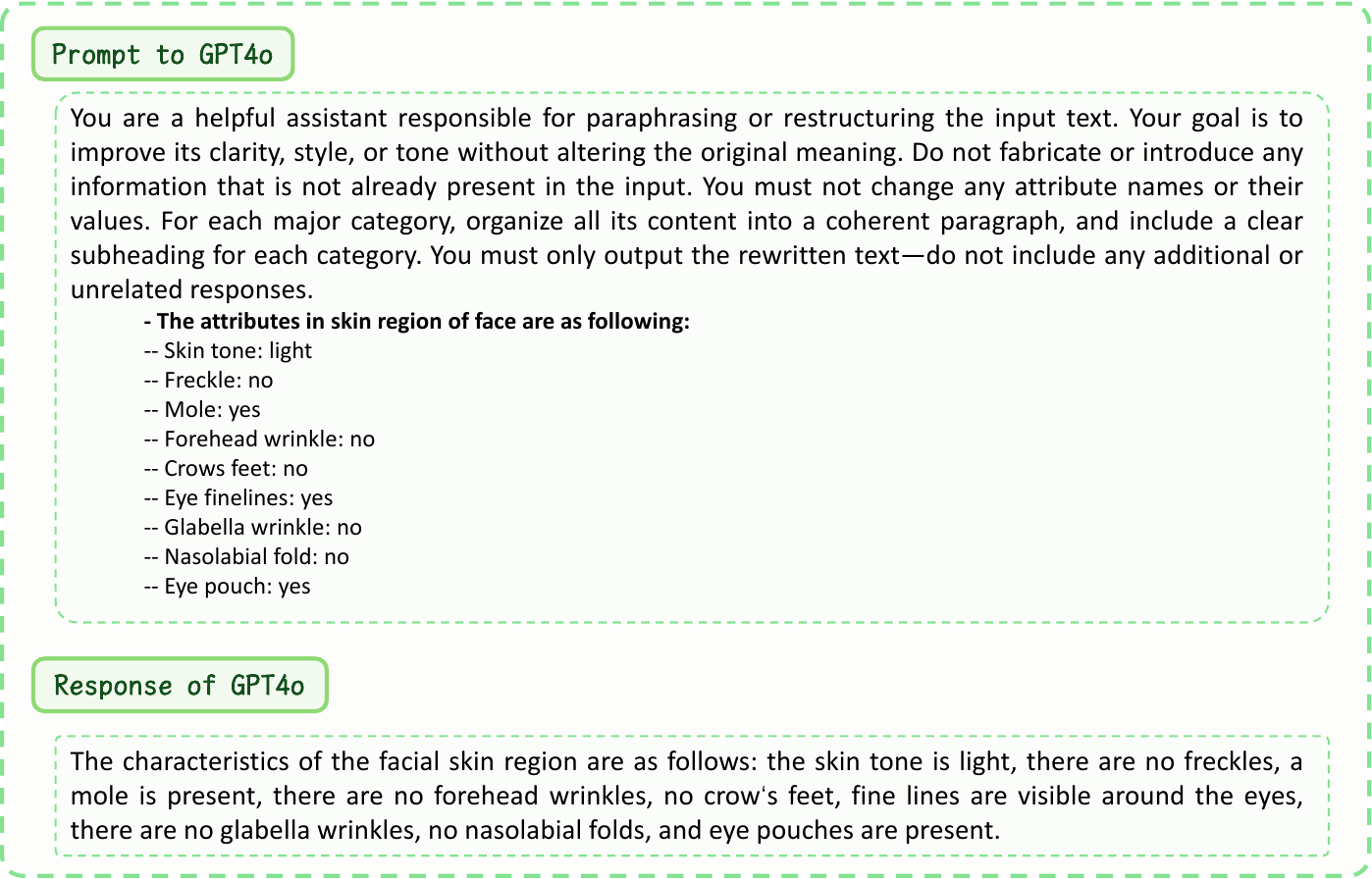}
  \caption{An example of using GPT-4o to generate a coherent textual description by reorganizing structured facial attribute data. This case focuses on attributes related to the skin region as a representative example.}
  \label{fig:gpt4o}
\end{figure}

\begin{figure}
  \centering
  \includegraphics[width=1\linewidth]{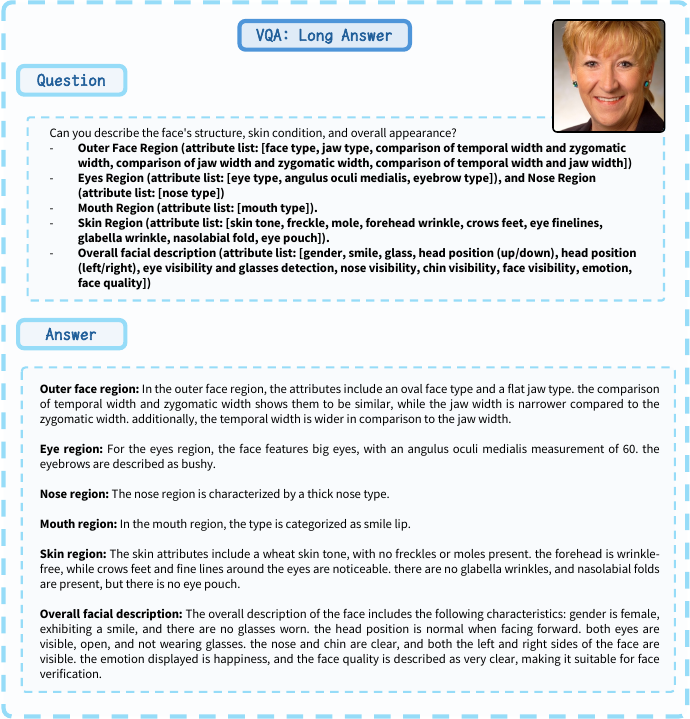}
  \caption{An example of long-answer VQA data. The images come from LAION-Face ~\citep{laion_face}}
  \label{fig:long_answer}
\end{figure}

To improve the ability of facial understanding of MLLMs, we proposed Facial Attributes Description Dataset $\mathcal{D}_{FA}$,  a multimodal dataset composed of \textit{high-resolution} facial images paired with rich \textit{facial attribute descriptions}. 
\paragraph{Data Collection\&Preprocessing}
High-resolution facial images are essential for effective facial understanding, as they preserve rich and detailed information. To this end, we collected a large number of high-quality facial images (with resolution more than $1024\times 1024$) from LAION-Face~\cite{laion_face}, filtering out those that were blurry or occluded. Specifically, we employed MTCNN~\cite{mtcnn} to crop faces from the raw images. We retained only those samples where facial landmarks could be reliably detected to ensure clarity. 
Finally, we selected faces with yaw and pitch angles below $15^\circ$ to ensure frontal and unobstructed views by 3DDFA~\cite{3dffa}.
\paragraph{Construction of VQA data}
We adopt a commercial facial analysis tool, MegVii’s official API\footnote{https://www.faceplusplus.com.cn/}, to obtain detailed facial 
attributes.
The facial attributes are shown in Figure \ref{fig:FA}. 
Subsequently, this structural information was used to build a VQA dataset that textually describes facial images. 
Specifically, we constructed three types of VQA data: multiple choice, short answer, and long answer.
For the multiple-choice and short-answer formats, we directly constructed the VQA data using the extracted facial attributes, as illustrated in Figure~\ref{fig:choice}.
In addition, to generate long-answer VQA data, we employed GPT-4o~\cite{gpt4o} to reorganize the structured facial attribute information into coherent and natural language descriptions, as demonstrated in Figure~\ref{fig:gpt4o}.
Figure~\ref{fig:long_answer} illustrates an example of the long answer format in $\mathcal{D}_{FA}$.
The naive MLLM will gain improved facial understanding abilities after training on the proposed facial VQA datasets.

\clearpage

\subsection{ID Discrimination Dataset}
\begin{figure}[h]
    \centering
    \includegraphics[width=1\linewidth]{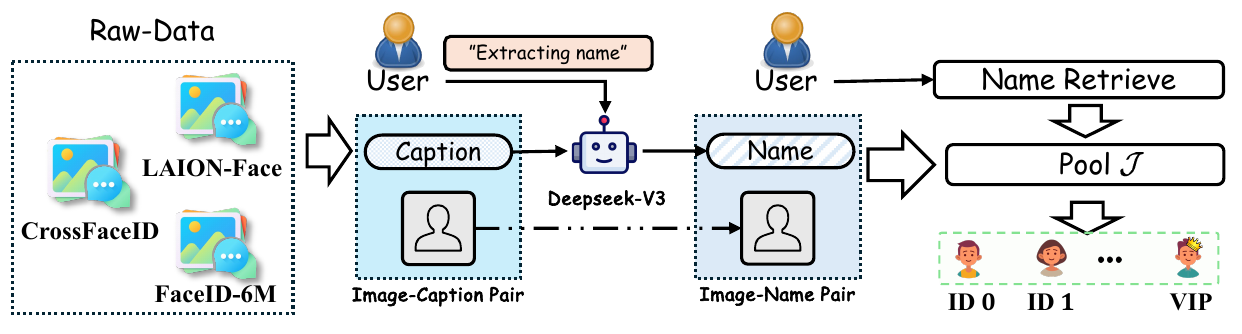}
    \caption{Pipeline for constructing the image-name pair pool $\mathcal{J}$.}
    \label{fig:imgj}
\end{figure}

\begin{figure}
    \centering
    \includegraphics[width=0.9\linewidth]{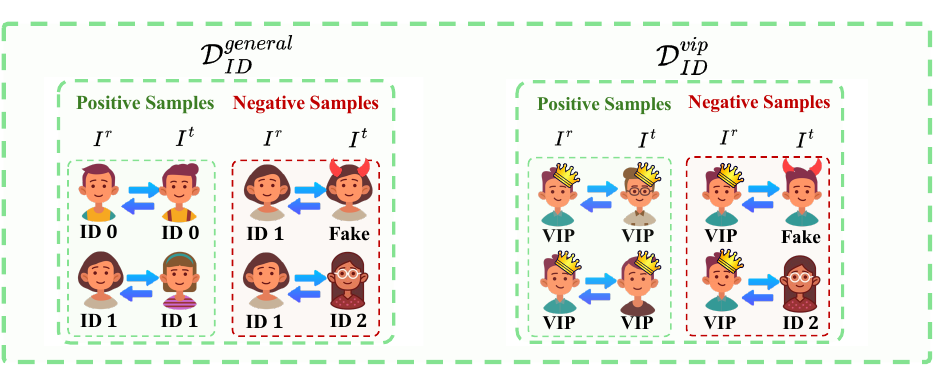}
    \caption{The composition of facial pairs in $\mathcal{D}_{ID}^{general}$ and $\mathcal{D}_{ID}^{vip}$.}
    \label{fig:face_pair}
\end{figure}

In this work, we formulate the Deepfake detection task as a fine-grained face recognition problem tailored to specific users. To support this identity verification task, we introduce two datasets: $\mathcal{D}_{ID}^{general}$ and $\mathcal{D}_{ID}^{vip}$. The former consists of facial image pairs drawn from arbitrary identities, whereas the latter comprises facial image pairs constructed specifically around a particular user.
\paragraph{Data Collection\&Preprocessing}
\label{subsec:did}
To construct these datasets, we first require a large volume of facial images with known identities. To this end, we source relevant images from existing open source datasets, including LAION-Face~\cite{laion_face}, FaceID-6M~\cite{faceid6m}, and CrossFaceID~\cite{crossfaceid}. 
As illustrated in Figure~\ref{fig:imgj}, we utilized the image-caption pairs from these open source datasets and employed Deepseek to extract names from the captions, thereby generating a pool $\mathcal{J}$ of image-name pairs.
We retrieved multiple images for each identity based on the extracted names and applied the same image pre-processing techniques as described in Section~\ref{subsec:favqa}. 
Subsequently, as illustrated in Figure~\ref{fig:face_pair}, we constructed facial image pairs for both $\mathcal{D}_{ID}^{general}$ and $\mathcal{D}_{ID}^{vip}$. The collected images were organized into a large number of pairs, each comprising a reference image $I^r$ and a test image $I^t$.
For $\mathcal{D}_{ID}^{general}$, positive pairs comprise two real images of the same identity (denoted as $I^r_{real}$–$I^t_{real}$–Same ID), while negative pairs include either two real images from different identities ($I^r_{real}$–$I^t_{real}$–Diff ID) or a real image paired with its corresponding fake counterpart ($I^r_{real}$–$I^t_{fake}$–Same ID).
To generate fake images, we employed SimSwap~\citep{simswap} and Arc2Face~\citep{arc2face}. Notably, when using SimSwap, we replaced the identity vector typically required for face swapping with a random noise vector $\sigma \sim \mathcal{N}(0,1)$, enabling the substitution of the inner face while preserving the outer facial features. In addition, we constructed partially swapped images by applying different masks to the eyes, nose, mouth, and inner face regions.
The ratio of samples in $\mathcal{D}_{ID}^{general}$ is $$(I^r_{real}\text{–}I^t_{real}\text{–Same ID}) : (I^r_{real}\text{–}I^t_{real}\text{–Diff ID}) : (I^r_{real}\text{–}I^t_{fake}\text{–Same ID})=2:1:1,$$
which ensures a balanced number of positive and negative samples.

Similar to $\mathcal{D}_{ID}^{general}$, the construction pipeline for $\mathcal{D}{ID}^{vip}$ followed a nearly identical procedure, with the key distinction that all reference images $I^r$ exclusively belonged to VIP users. Due to the few number of the real images of VIP users, the ratio was adjusted to $$(I^r_{real}\text{–}I^t_{real}\text{–Same ID}) : (I^r_{real}\text{–}I^t_{real}\text{–Diff ID}) : (I^r_{real}\text{–}I^t_{fake}\text{–Same ID})=1:5:5,$$
where the number of negative samples is ten times that of the positive samples.
% High-resolution facial images are essential for effective facial understanding, as they preserve rich and detailed information. To this end, we collected a large number of high-quality facial images (with resolution more than $1024\times 1024$) from LAION-Face~\cite{laion_face}, filtering out those that were blurry or occluded. Specifically, we employed MTCNN~\cite{mtcnn} to crop faces from the raw images. We retained only those samples where facial landmarks could be reliably detected to ensure clarity. 
% Finally, we selected faces with yaw and pitch angles below $15^\circ$ to ensure frontal and unobstructed views by 3DDFA~\cite{3dffa}.


\begin{figure}
    \centering
    \includegraphics[width=0.9\linewidth]{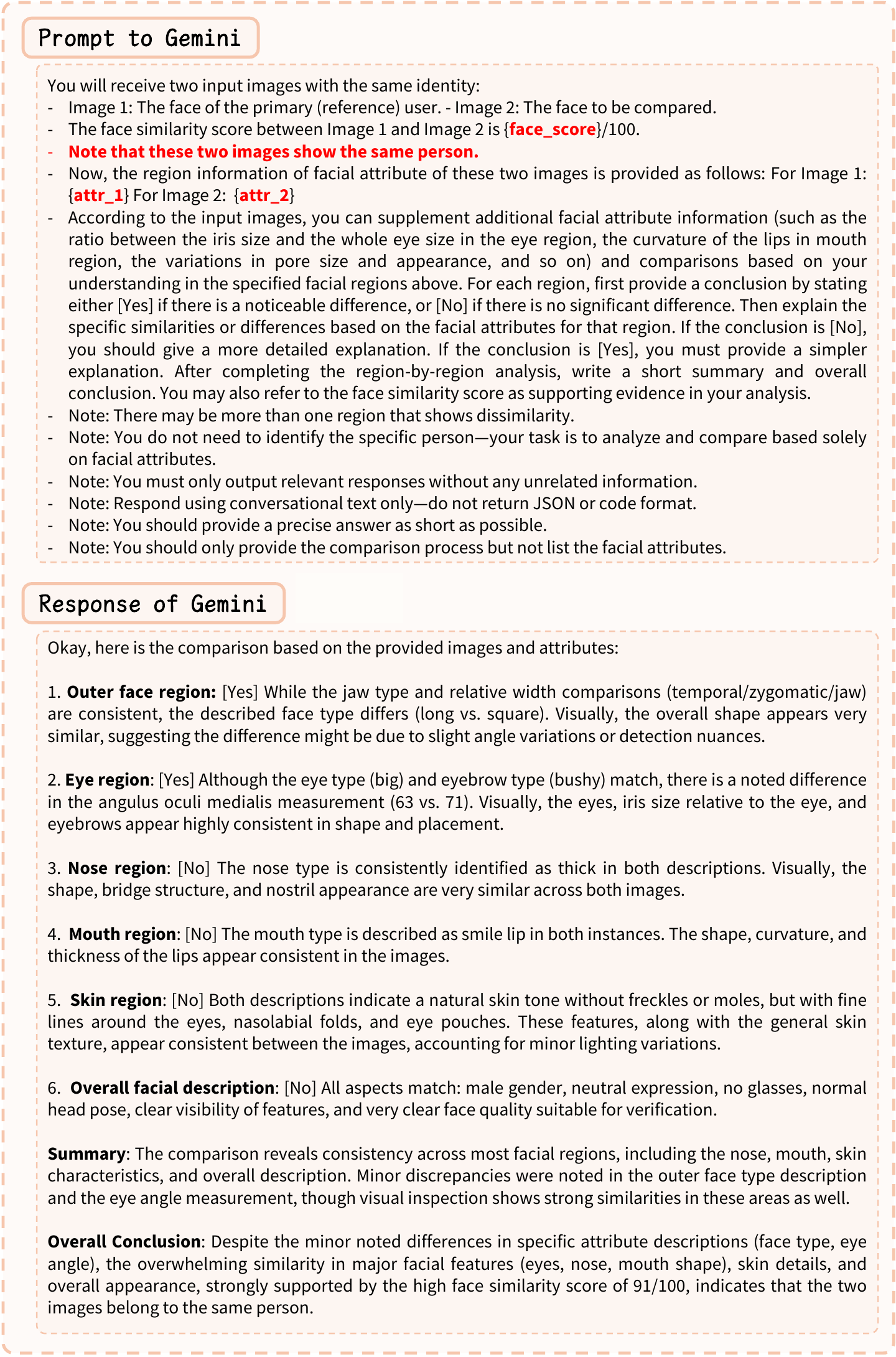}
    \caption{An instance of using Gemini to generate a textual description for facial discrimination. \texttt{face score} is the face similarity score calculated by face recognition models. \texttt{attr 1} and \texttt{attr 2}, formatted as shown in the answer in Figure~\ref{fig:long_answer}, should be filled with attribute information generated by the captioning model.}
    \label{fig:gemini}
\end{figure}

\paragraph{Construction of VQA data}
After constructing the facial image pair dataset, we first annotate facial attributes using a captioning model (Qwen-2.5-VL-7B), which was fine-tuned on the $\mathcal{D}_{FA}$ dataset. These attribute annotations, together with facial similarity scores, are then provided as input to Gemini\footnote{Gemini API version in use: 2.5-pro-exp-03-25.}~\cite{Gemini}, which is tasked with analyzing the similarities and differences between the two faces. Figure ~\ref{fig:gemini} illustrates an example of the prompt used during this process.
The prompts are adjusted based on the sample type. For positive samples, Gemini is explicitly instructed with the note: “\texttt{Note that these two images show the same person.}” For negative samples, the prompt states: “\texttt{Note that these two images show different persons.}” Furthermore, in cases where the negative sample involves a fake image, we provide a more nuanced prompt: “\texttt{Although the faces may appear similar, they are not the same person.}”
Importantly, Gemini was constrained to base its reasoning strictly on the provided attribute annotations, thereby mitigating hallucinations.
Finally, the VQA data and the corresponding facial image pairs constituted our dataset $\mathcal{D}_{ID}^{general}$ and $\mathcal{D}_{ID}^{vip}$

\subsection{VIPEval: User-Specific Evaluation Dataset }
\begin{figure}
    \centering
    \includegraphics[width=0.9\linewidth]{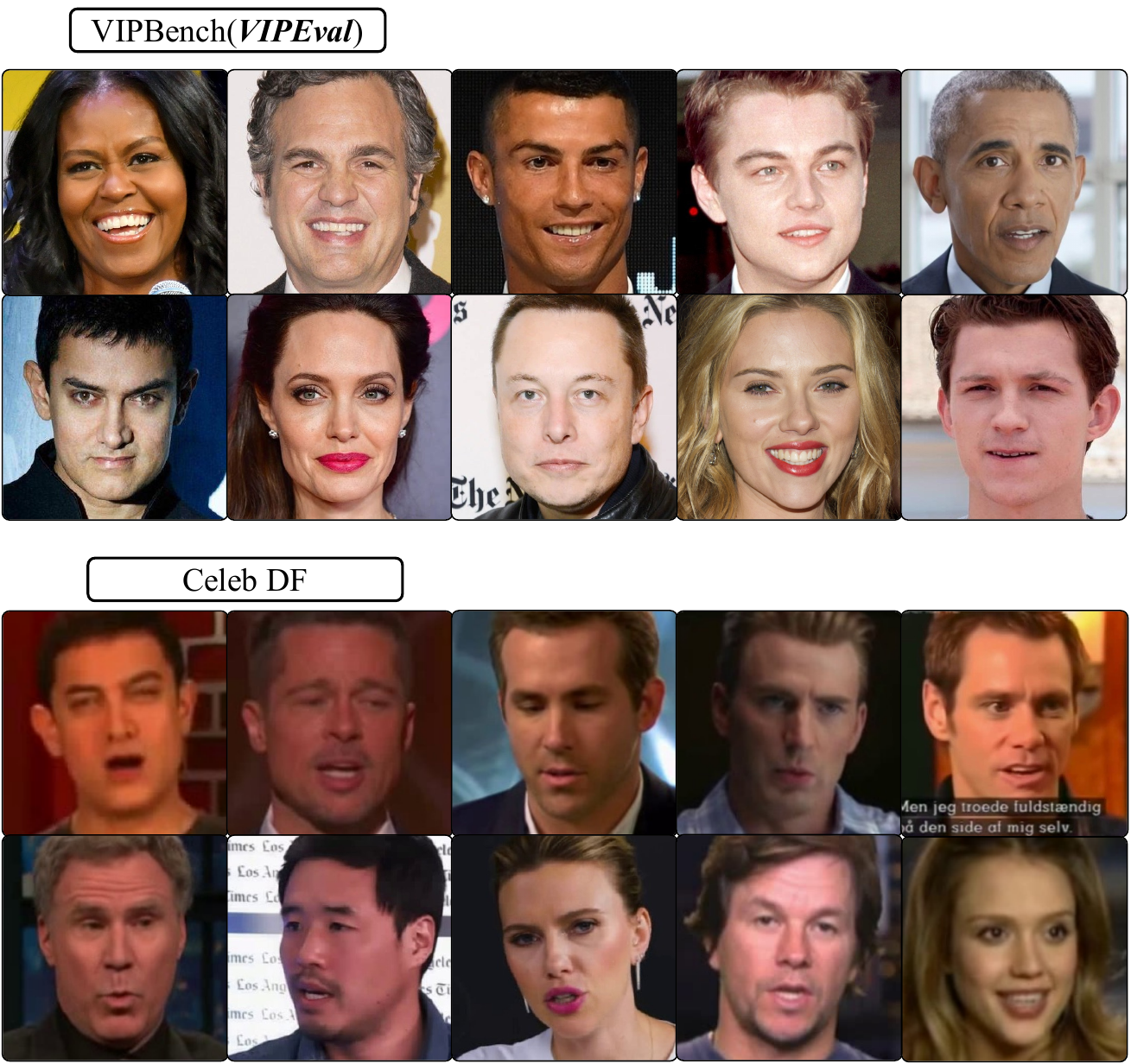}
    \caption{Comparasion of VIPBench (Ours) and CelebDF~\cite{li2019celeb}. The quality of authentic images is higher in VIPBench. The images in VIPBench are from LAION-Face~\citep{laion_face}, FaceID-6M~\citep{faceid6m}, and CrossFaceID~\cite{crossfaceid}.}
    \label{fig:real_img}
\end{figure}

To better evaluate the effectiveness of user-specific protection methods and ID-aware detectors, we constructed a multi-resolution dataset that includes seven types of face swapping and seven types of fully synthetic face forgeries. In this section, we present representative samples from the dataset. 
In Figure~\ref{fig:real_img}, we present a comparison between authentic images from CelebDF~\cite{li2019celeb} and our VIPEval dataset. VIPEval includes a large number of images with various resolutions for each identity that capture significantly more facial detail.

\paragraph{Data Collection\&Preprocessing}
The data collection and pre-processing procedure in VIPEval follows the same steps as described in Section~\ref{subsec:favqa}. Building on this, we collected images for 22 target identities, obtaining approximately 40–60 real images per identity. For each identity, 20 real images were selected and fixed as the real samples in the test set, as well as the target images for generating fake content. The remaining 20–40 images were used as training data for that identity, and were used to construct $\mathcal{D}_{ID}^{vip}$ following the method described in Section~\ref{subsec:did}.
The training set $\mathcal{D}_{ID}^{vip}$ is individually customized for each identity.

\newpage
\clearpage

\end{document}